\documentclass[lettersize,journal]{IEEEtran}
\usepackage{amsmath,amsfonts}
\usepackage{algorithmic}
\usepackage{algorithm}
\usepackage{array}
\usepackage{textcomp}
\usepackage{stfloats}
\usepackage{url}
\usepackage{verbatim}
\usepackage{graphicx}
\usepackage{cite}
\usepackage{xspace}

\usepackage{tcolorbox} 
\usepackage{xcolor} 
\usepackage{enumitem} 
\usepackage{mdframed} 

\newcommand\lwg[1]{\sethlcolor{orange} \hl{lwg: #1}}
\newcommand{\yrj}[1]{{\color{red}{#1}}}
\newcommand{\mwx}[1]{#1}
\newcommand{\sys}{\texttt{EdgeMoE}\xspace}

\usepackage{wrapfig}
\usepackage{comment}
\usepackage{colortbl}

\usepackage{hyperref}

\usepackage{color,soul}
\usepackage{graphics}
\usepackage{hyperref}
\usepackage{lipsum}
\usepackage{tikz}
\usepackage{enumitem}	
\usepackage{listings} 

\usepackage{booktabs}	
\usepackage{lipsum}

\usepackage{capt-of}	

\usepackage{todonotes}  
\usepackage{subcaption} 

\definecolor{refkey}{rgb}{0,0,1}
\definecolor{labelkey}{rgb}{0,0,1}

\usepackage{cleveref}
\AtBeginDocument{\DeclareCaptionSubType{lstlisting}}
\crefname{sublstlisting}{listing}{listings}
\Crefname{sublstlisting}{Listing}{Listings}

\usepackage{mathrsfs}	
\usepackage{amsfonts}

\usepackage[export]{adjustbox} 

\usepackage{boldline}					

\usepackage{multirow}

\renewcommand{\paragraph}[1]{\vskip 3pt\noindent\textbf{#1 }}	 

  {\begin{list}{$\bullet$}%
     {\setlength{\parsep}{0pt}%
      \setlength{\topsep}{0pt}%
      \setlength{\itemsep}{2pt}}}%
  {\end{list}}
\newcommand\Note[1]{\sethlcolor{yellow} \hl{#1}} 
\newcommand\Noted[1]{} 

\definecolor{darkblue}{rgb}{0.0, 0.0, 0.55}
\definecolor{mygreen}{HTML}{ADFF2F}
\definecolor{mylightgray}{gray}{0.8}

\newenvironment{myitemize}%
  {\begin{itemize}
	[leftmargin=0cm,
		itemindent=.3cm,
		labelwidth=\itemindent,
		labelsep=0pt,
		parsep=0.0pt,
		topsep=0.5pt,
		itemsep=0.5pt,
		align=left]
  }%
  {\end{itemize}}    

\newenvironment{myenumerate}%
  {\begin{enumerate}
	[leftmargin=.cm,itemindent=.5cm,labelwidth=\itemindent,
		labelsep=0pt,
		parsep=1pt,
		topsep=1pt,
		itemsep=3pt,
		align=left]
  }%
  {\end{enumerate}}    

\newcommand{\tfm}{Transformer\xspace}

\makeatletter
\def\@copyrightspace{\relax}
\makeatother

\begin{document}

\title{\sys: Empowering Sparse Large Language Models on Mobile Devices}


\author{
\IEEEauthorblockN{
Rongjie Yi\IEEEauthorrefmark{1},
Liwei Guo\IEEEauthorrefmark{2},
Shiyun Wei\IEEEauthorrefmark{3}, 
Ao Zhou\IEEEauthorrefmark{1}, 
Shangguang Wang\IEEEauthorrefmark{1}~\IEEEmembership{Senior Member,~IEEE}, and
Mengwei Xu\IEEEauthorrefmark{1}~\IEEEmembership{Member,~IEEE},
}
\IEEEauthorblockA{\IEEEauthorrefmark{1}Beijing University of Posts and Telecommunications, China \\}
\IEEEauthorblockA{\IEEEauthorrefmark{2}University of Electronic Science and Technology of China, China \\}
\IEEEauthorblockA{\IEEEauthorrefmark{3}Zhongguangcun Laboratory, China}

}

\markboth{ IEEE TRANSACTIONS ON MOBILE COMPUTING}%
{Shell \MakeLowercase{\textit{et al.}}: A Sample Article Using IEEEtran.cls for IEEE Journals}

\IEEEpubid{0000--0000/00\$00.00~\copyright~2021 IEEE}

\maketitle

\begin{abstract}
Large language models (LLMs) such as GPTs and Mixtral-8x7B have revolutionized machine intelligence due to their exceptional abilities in generic ML tasks. 
Transiting LLMs from datacenters to edge devices brings benefits like better privacy and availability, but is challenged by their massive parameter size and thus unbearable runtime costs.

To this end, we present \sys, an on-device inference engine for mixture-of-expert (MoE) LLMs -- a popular form of sparse LLM that scales its parameter size with almost constant computing complexity. 
\sys achieves both memory- and compute-efficiency by partitioning the model into the storage hierarchy:
non-expert weights are held in device memory;
while expert weights are held on external storage and fetched to memory only when activated.
This design is motivated by a key observation that expert weights are bulky but infrequently used due to sparse activation.
To further reduce the expert I/O swapping overhead,
\sys incorporates two novel techniques: 
(1) expert-wise bitwidth adaptation that reduces the expert sizes with tolerable accuracy loss;
(2) expert preloading that predicts the activated experts ahead of time and preloads it with the compute-I/O pipeline. 
On popular MoE LLMs and edge devices, \sys showcase significant memory savings and speedup over competitive baselines.
The code is available at \url{https://github.com/UbiquitousLearning/mllm}.

\end{abstract}



\section{Introduction}\label{sec:intro}
\IEEEPARstart{L}{arge} language models (LLMs), e.g., GPTs~\cite{radford2018improving,radford2019language,brown2020language,ouyang2022training,openai2023gpt4techinicalreport,eloundou2023gpts} and LLaMa~\cite{touvron2023llama,touvron2023llama2}, are reshaping machine intelligence for their remarkable performance on generic NLP tasks, few-shot ability, and scalability.
While born in datacenter warehouses, LLMs are gradually sinking to edge devices like personal PCs, smartphones, and even IoTs, for better data privacy, availability, and personalization.
In this trend, LLMs not only greatly advance the state-of-the-art performance of edge ML tasks compared to traditional DNNs, but also enable many new, exciting edge applications~\cite{chatgpt-googleplay}.
For instance, Qualcomm has deployed a text-to-image generative LLM model with more than 1 billion parameters entirely on smartphones~\cite{qualcommstabledef}.
Huawei has embedded a multimodal LLM into its smartphones to facilitate accurate natural language-based content searching~\cite{huaweuai}.

Landing LLMs on mobile devices face a key challenge of its vast parameter size and consequently unaffordable runtime cost.
To alleviate this issue, mixture-of-experts (MoE) architecture~\cite{jacobs1991textordfeminineadaptive,fedus2022review}, which allows only part of the LLM to be activated in per-token decoding, has been proposed recently.
The most representative MoE design is to substitute the single feed-forward network (FFN) within each transformer block with many experts (each is an independent FFN).
During inference, a trainable function (namely \textit{router}) within each transformer block routes the input to only Top-K (K=1 or 2) of all experts.
More design details of MoE LLMs are presented in $\S$\ref{sec:bkgnd}.
Such MoE-based LLMs have been extensively researched~\cite{shazeer2017outrageously,lepikhin2020gshard,fedus2022switch,du2022glam,lewis2021base,roller2021hash,zhou2022mixture} and adopted in industry~\cite{microsoft-moe}.
Intuitively, the sparse activation makes lots of sense as LLMs go larger and serve as a foundation model for various tasks, different tasks or input data could require only a tiny, different portion of the model to work, just as how human brains function~\cite{friston2008hierarchical}.

\textbf{Pros and Cons of MoE}
Through their sparsity design, MoE LLMs can scale their parameter size and ability with almost constant computing complexity, making them a good fit for edge devices whose storage is much more cost-efficient and scalable than the computing capacity.
Empirically, to hold a 105B-parameter GLaM~\cite{du2022glam}, a device needs 1T (external) storage that costs only less than \$100;
yet to execute it at a reasonable speed, e.g., 10 tokens/sec, five high-end GPUs are demanded that cost about \$65K.
However, MoE LLMs are too large to fit into device memory (detailed in $\S$\ref{sec:back:exps}).
Simply scaling down the expert number could significantly degrade the model performance~\cite{fedus2022switch};
either, frequently swapping weights between memory and storage incurs huge overhead due to the autoregressive nature of LLM.


\textbf{\sys: an expert-centric LLM engine for mobile devices.}
This work presents \sys, the first on-device LLM inference engine that can scale out the model size (expert number) with both memory and time efficiency.
The overall design of \sys is based on a unique observation:
most computations reside in a small portion of non-expert weights (``hot weights'') that can be held in device memory;
while the expert weights contribute most of the memory footprint but only a tiny part of computations (``cold weights'').
\IEEEpubidadjcol
Thereby, \sys differentiates the positions of experts and non-experts in the storage hierarchy (faster RAM vs. larger Disk).
Specifically, it permanently hosts all hot weights in memory since they are used per-token inference; while the rest of the memory budget is used as an \textit{expert buffer} for the cold expert weights.

With the expert buffer design, \sys only needs to load the activated experts on demand from storage to memory.
However, this I/O overhead is still significant as compared to processing, e.g., up to 4.1$\times$ delay on Jetson TX2, as will be shown in $\S$\ref{sec:back:moe}.
To address this issue, there are two general approaches:
one is to directly reduce the I/O data, e.g., by quantizations~\cite{frantar2022gptq,xiao2023smoothquant,lin2023awq,yuan2023rptq,liu2023llm,chee2023quip};
another one is to pipeline the I/O with computing to hide its latency~\cite{guo2023sti}.
While both directions have been well studied, adopting them in MoE models faces unique challenges:
(1) 
Sophisticated quantization algorithms~\cite{zadeh2020gobo,aji2020compressing,kim2023squeezellm} achieve higher compression ratio, yet incurring significant pre-processing time for deserialization and decompression as discussed in $\S$\ref{sec:design:quantization}. 
More vanilla quantizations~\cite{han2015deep}, on the other hand, cannot effectively reduce the expert's I/O.
(2) Unlike static models that have a fixed pre-determined execution pattern, experts in MoE are dynamically activated and the system cannot derive a priori knowledge until the router functions.
As such, there is no room for \sys to pre-load a to-be-activated expert.
Disregarding the knowledge of expert activation, one might simply cache the experts that are more frequently activated to increase the expert hit ratio;
however, this approach brings limited benefit since the activation frequency across experts is purposely trained to be balanced~\cite{fedus2022review}.

In response, \sys proposes two novel designs.

\textbf{Expert-wise bitwidth adaptation.}
\sys augments a preprocessing-lightweight quantization method,  per-channel linear quantization~\cite{kim2022mixture}, with expert-level bitwidth adaptation.
It is based on a crucial observation that experts across different layers or even in the same layer exhibit different impacts on model accuracy after being quantized.
Therefore, \sys employs a fine-grained, expert-wise bitwidth adaptation to fully leverage the model redundancy.
At offline, \sys progressively lower the bitwidth of a few experts that are most robust to quantization, till the accuracy degradation meets a tolerable threshold specified by users.
The selection of which experts to further quantize also jointly considers how much the lower-bitwidth quantization could boost inference speed.
Ultimately, \sys obtains a mixed-precision model that achieves the target accuracy with the smallest possible model size, i.e., the fastest loading time. 

\textbf{In-memory expert management.}
To enable the I/O-compute pipeline, \sys predicts which expert will be activated before its router functions.
The design is motivated by a novel observation: the expert activation paths (i.e., the set of sequentially activated experts per token) taken in practice are highly unbalanced and skewed.
It indicates significant correlations between expert activations, as further confirmed by our experiments in $\S$\ref{sec:design:mm:act}.
Therefore, during the offline phase, \sys builds a statistic model to estimate the probability of expert activation in the current layer based on the activations of previous layers. 
In online inference, \sys queries this model and preloads the most possible expert ahead of activation for the I/O-compute pipeline. 
Additionally, \sys designs a novel cache eviction policy for the expert buffer, leveraging both the activation frequency and their relative positions to the current execution.
Overall, both the predict-then-preload and the eviction techniques maximize the expert cache hit ratio when they are activated.

\textbf{Results} 
We’ve implemented a prototype of \sys atop PyTorch that fully realizes the above techniques.
It takes a memory budget and a tolerable accuracy loss as input from developers and automatically optimizes execution latency.
We then perform extensive experiments to evaluate \sys's performance through 7 MoE-based LLMs and 2 embedded platforms including Raspberry Pi 4B (CPU) and Jetson TX2 (GPU).
Compared to holding the whole model in device memory, \sys reduces memory footprint by 1.05$\times$--1.18$\times$;
compared to memory-optimized baselines such as dynamically loading expert and STI~\cite{guo2023sti},
\sys achieves 1.19$\times$--2.77$\times$ inference speedup.
For the first time, \sys enables fast inference for $>$10B-sized LLMs on COTS edge devices like Jetson TX2 with negligible accuracy loss ($\le$2\%).
The ablation study further shows that each individual technique of \sys contributes to significant improvements.

\textbf{Contributions}
The paper makes following contributions:
\begin{itemize}[leftmargin=*,topsep=1pt]
	\item We perform preliminary experiments to demystify the performance of MoE LLMs on edge devices and analyze the implications.
	\item We present \sys, an on-device MoE engine with one key design that treats memory as a cache for experts that are held in external storage when not activated.
	\item We further incorporate two novel techniques, namely expert-wise bitwidth adaptation and in-memory expert management, to reduce the expert I/O overhead of \sys.
	\item We demonstrate the effectiveness of \sys through extensive experiments.
\end{itemize}

\section{Pilot Experiments and Analysis} \label{sec:bkgnd}

%
%
%
%
%

\subsection{A Primer on LLM with Mixture-of-Experts}\label{sec:back:moe}

\begin{figure*}[t]
	\centering					
	\includegraphics[width=0.98\textwidth]{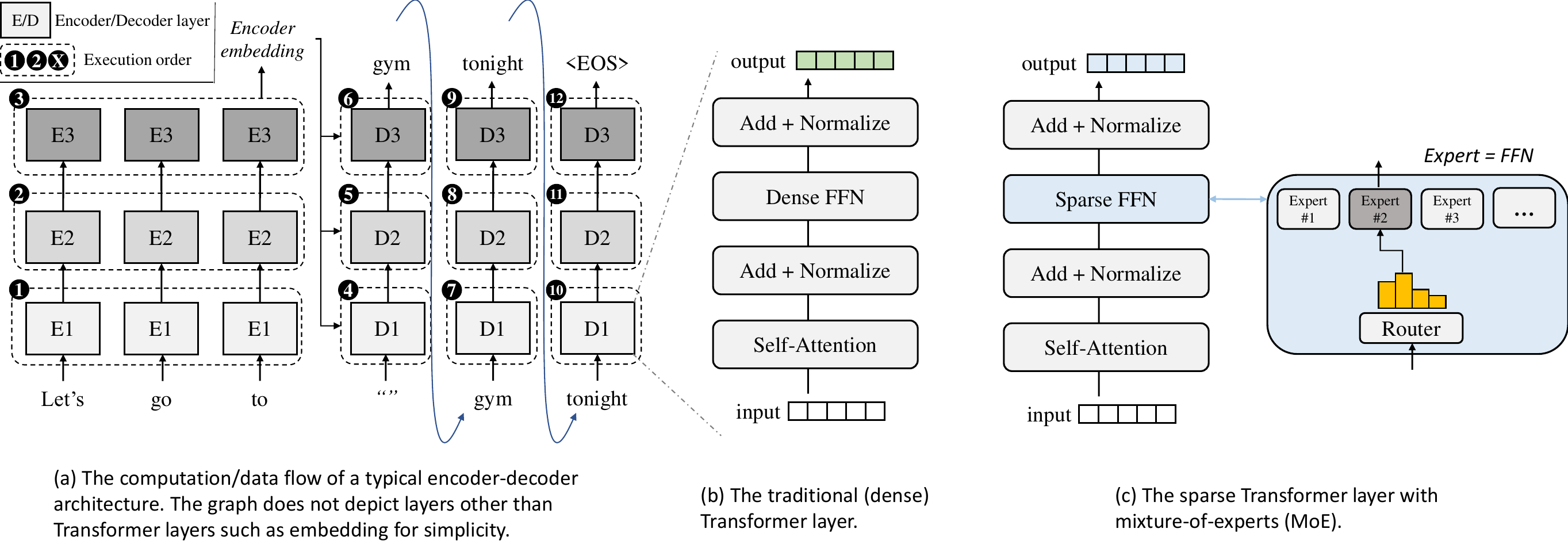}
	\caption{Illustrations for the inference procedure of a typical encoder-decoder language model, as well as the dense and sparse MoE-enabled Transformer layer architecture. In (a): The Transformer layers are executed in the order denoted by the numbers in the black circles, and the nodes that use the same set of model parameters (i.e., nodes representing the same layer) are filled with the same color.
	}
	\label{fig:MoE}
\end{figure*}

This work focuses on encoder-decoder\footnote{Decoder-only LLMs like GPTs~\cite{radford2018improving,radford2019language,brown2020language,ouyang2022training,openai2023gpt4techinicalreport,eloundou2023gpts} can be treated as a special case of encoder-decoder therefore is also supported by our system.}, one of the most popular LLM architectures nowadays.
The encoder processes the input sequence and compresses this information into a continuous intermediate representation, while the decoder takes this representation and generates (predicts) an output sequence.
A unique characteristic of the decoder is that it generates tokens in an \textit{autoregressive} manner, i.e., appending the last output token to the end of the input sequence when generating the next token (token-wise dependency).
Figure~\ref{fig:MoE}(a) illustrates a simplified computation and dataflow graph of the LLM inference process with three Transformer layers.
Both encoder and decoder are underpinned by \tfm{} layers~\cite{vaswani2017attention}, each consisting of a set of attention heads (for extracting word-pair relationships), FFNs (for processing and enhancing information representation with non-linearity), and other minor operators, as shown in Figure~\ref{fig:MoE}(b).

A recent trend is to deploy \textit{sparse} FFNs -- a set of ``experts'' which is selected at runtime via small-sized, offline-trained ``routers'', as illustrated in Figure~\ref{fig:MoE}(c). 
As a result, MoE architecture can scale the model parameter size with sublinearly increased computing complexity.
This is because only a fixed set (typically 1 or 2) of experts/weights will be activated for each token.
For instance, GLaM~\cite{du2022glam}, an MoE-based LLM, achieves considerably higher accuracy on NLP tasks than GPT-3 with only half of its computation cost.  Mixtral-8x7B~\cite{jiang2024mixtral} also reports comparable performance  to GPT-3.5 at only 10\% runtime cost.
Such parameter scalability makes MoE-based LLMs good candidates for edge devices in terms of their constrained computing capacity.

\subsection{On-device Sparse LLM Inference}\label{sec:back:exps}

\begin{figure}[t]
	\centering
	\subfloat[Test Loss]{
		\includegraphics[width=0.23\textwidth]{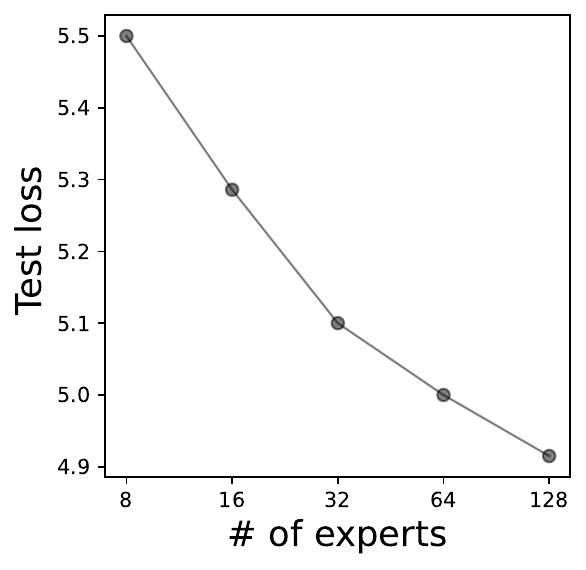}%
	}
	\subfloat[peak memory usage]{
		\includegraphics[width=0.23\textwidth]{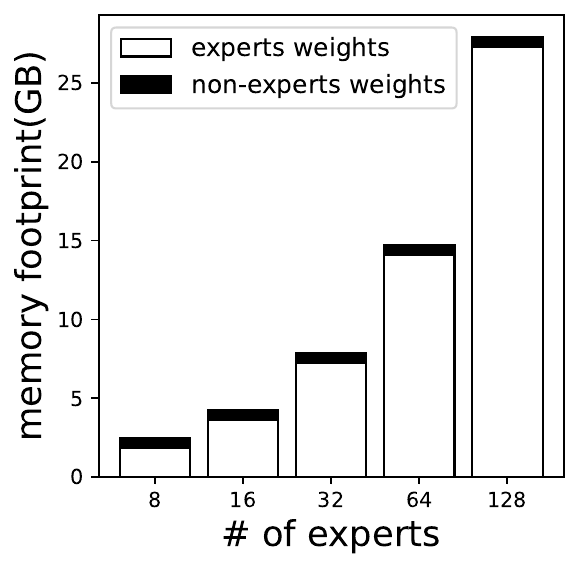}%
	}
	\caption{
		The importance and cost in scaling up expert number.
		(left): with more experts per layer, the model accuracy continuously improves. Dataset: C4~\cite{huggingface-c4}.
		Numbers are from Switch Transformers paper~\cite{fedus2022switch}.
		(right): with more experts per layer, the peak memory usage increases almost linearly.
    }
	\label{fig:bkgnd-modelmem}
\end{figure}
\begin{figure}[!t]
	\centering
	\subfloat[Jetson TX2]{
		\includegraphics[width=0.23\textwidth]{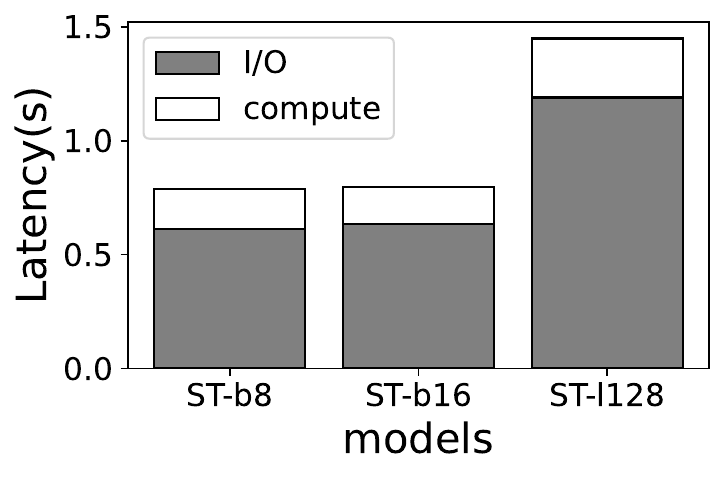}
	}
	\subfloat[Raspberry Pi 4B]{
		\includegraphics[width=0.23\textwidth]{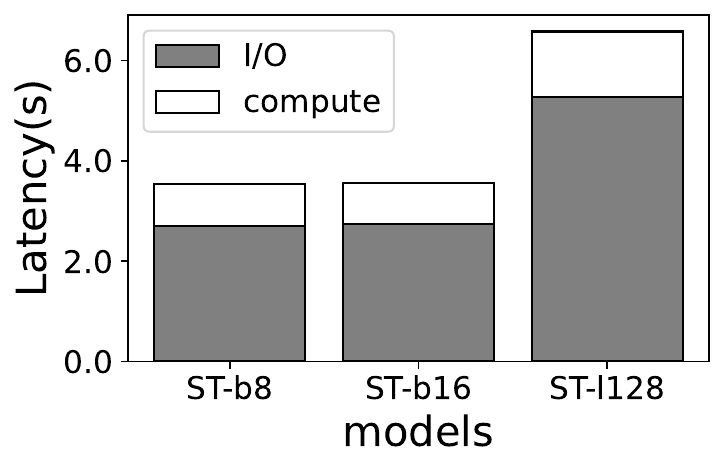}
	}
	\caption{
		Per-token decoder's inference time
	}
	\label{fig:motivation-decinfertime}
\end{figure}

With great sparsity comes great model size.
Sparsely scaled-out LLMs stress memory, the key resource on an edge device. 
To better understand their implications to an edge device, especially the memory/computation tradeoff and execution characteristics, we characterize the execution of Switch Transformer~\cite{fedus2022switch}(abbreviated as ST), one of the most popular MoE-based sparse LLMs by Google, on two Commercial Off-The-Shelf(COTS) SoCs Jetson TX2 and Raspberry Pi 4B.
We make the following crucial observations as follows.


\noindent \textbf{(1) Expert weights bloat device memory.}
While improving the model accuracy, the expert weights quickly bloat the model sizes as their number increases.
Google has shown that by scaling up the expert number per FFN from 8 to 256, the model capacity continuously and remarkably improves~\cite{fedus2022switch}.
However, as shown in Figure~\ref{fig:bkgnd-modelmem}, the increased number of experts leads to a huge peak memory footprint that is unaffordable by edge devices.
For instance, Raspberry Pi 4B with 8GB memory can only hold the smallest Switch Transformers variant with 8 experts per FFN.
Consequently, the memory wall severely limits the scalability of MoE-based LLMs, which is crucial to its success.
Note that even if the device memory is large enough (e.g. Jetson TX2 with 8GB RAM) to hold a whole model in memory, the large model size makes it a likely victim of OS memory management. 
The result is the model only can serve a few inferences before getting recycled. 

One might resort to a layer-by-layer swapping strategy~\cite{guo2023sti} to handle memory inefficacy.
However, due to the auto-regressive nature of LLMs, the whole model weights need to be loaded for decoding each token.
As a result, the I/O loading time could be 30.9$\times$ more than computing, making the inference extremely slow.
	

\noindent \textbf{(2) Experts weights are bulky but cold.}
For MoE-based \tfm models, the weight parameters of the expert networks in the MoE structure are called expert parameters, and the other parameters are non-expert parameters.  
We find that most computations during inference reside in a small portion of weights (non-experts), while most of the weights (experts) contribute to a small portion of computations.
This is attributed to the sparse activation nature of experts.
Taking Switch Transformers base-16(abbreviated as ST-base-16, 16 means each MoE layer contains 16 experts) as an instance, the experts contribute 86.5\% of total memory usage while only 26.4\% of the computation.
	
Intuitively, the above characteristics naturally fit the device storage hierarchy (faster RAM vs. larger Disk).
Therefore, we could use a \textit{discriminative swapping} strategy by holding all non-experts in memory but only swapping in/out experts between memory and disk.
Specifically, an expert is loaded into memory only when it is activated by the router; once used, its memory is released immediately.
In such a case, the memory usage could be as less as the size of all non-expert weights plus the size of one expert.
Meanwhile, the I/O overhead is reduced to one expert's weight per layer.

\begin{figure}[t]
	\centering
	\subfloat[ST-base-8]{
		\includegraphics[width=0.23\textwidth]{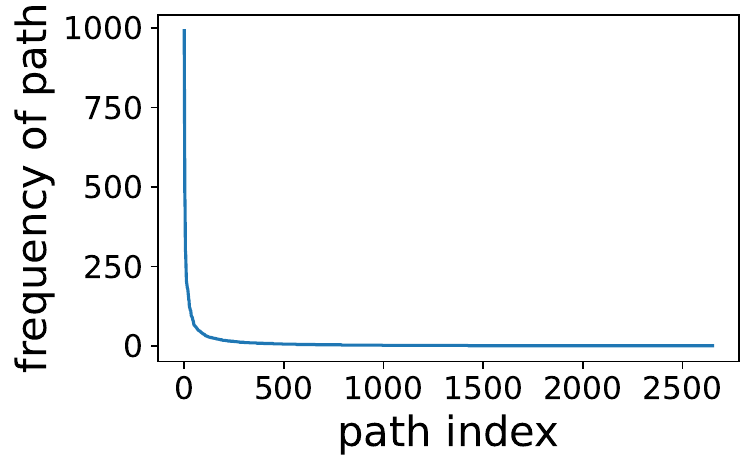}%
	}
	\subfloat[ST-base-16]{
		\includegraphics[width=0.23\textwidth]{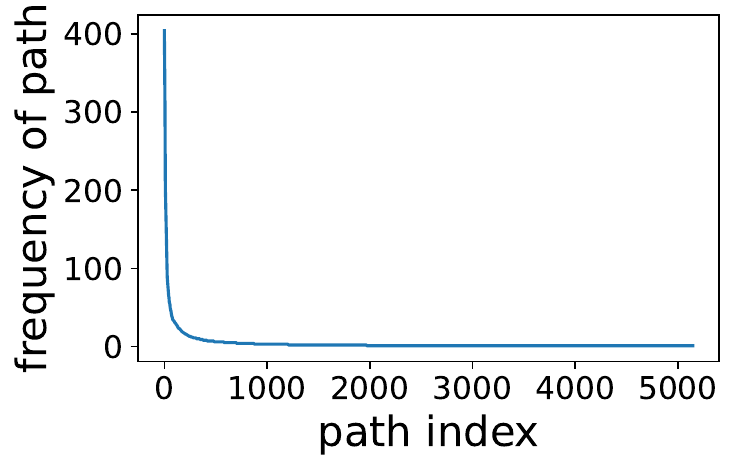}%
	}
	\caption{
		The distribution of expert activation paths obtained on training dataset follows power law.
		Path index: the expert's activation path sorted in descending order of frequency.
		Datssets: SAMsum, GLUE.
	}
	\label{fig:bkgnd-path}
\end{figure}	

\noindent \textbf{(3) Expert weight computation vs I/O asymmetry.} 
Unfortunately, even loading one expert per layer significantly degrades the execution performance of MoEs.
With the above discriminative swapping strategy, we find that the per-sample inference time (generating a whole sequence with multiple tokens) is slow on Jetson TX2 (e.g., more than 17secs on Jetson TX2 on average), and the decoding time dominates due to its autoregressive nature.
We further break down the latency to the compute (inference) and I/O (experts loading) in Figure~\ref{fig:motivation-decinfertime} and find that the latter contributes the most.
Compared to an oracle case with infinite memory (so no experts I/O), the per-token decoding time is increased by 3.2$\times$--3.9$\times$ and 3.3$\times$--3.8$\times$ on Jetson TX2 and Raspberry Pi 4B, respectively.

\noindent \textbf{(4) Compute/IO pipelining is hindered by the data dependency.}
One may further leverage the compute/IO parallelism and overlap the expert loading with weight computation, similar to STI~\cite{guo2023sti}.
However, we find such an approach unfeasible due to the following reasons.
\begin{itemize}[leftmargin=*,topsep=1pt]
\item \textit{Expert activation dependency.}
Unlike standard \tfm{} models targeted by STI, which sequentially preload layers, MoE \tfm{}s only decide \textit{which} experts to load when the prior layer finishes computation.
Such expert-level activation dependency prohibits the compute/IO pipeline.
\item \textit{Expert activation frequency.}
One might preload ``hot'' experts with a higher chance to be activated into memory as a pipeline, i.e., a frequency-based cache strategy.
However, we find such an approach not beneficial as the experts in the same layer have a similar chance to be activated, as demonstrated by our experiments depicted in Figure~\ref{fig:motivation-topk}(left).
Such a balanced activation phenomenon is not surprising, because the training mechanism designs it to be so to maximize the GPU usage at training time~\cite{fedus2022switch}.
\end{itemize}

\noindent \textbf{(5) Expert activation path follows power law distribution.}
While the overall activation frequency of each single expert is well balanced, we find that the activation path, i.e., a vector of the activated experts of all transformer layers for a token, is highly skewed.
Figure~\ref{fig:bkgnd-path} depicts the distribution of activation path obtained on two MoE models and two datasets.
It presents a power law distribution: the 20\% most frequently activated paths contribute to more than 99\% of the activation cases of all tokens.
This observation implies that the expert activation across different layers could be non-iid, which drives us to a deeper analysis of the expert activation correlation later in $\S$\ref{sec:design:cache}.

\section{\sys Design}\label{sec:design}

\subsection{Overview}


\paragraph{System model}
\sys is the first execution engine to enable \textit{fast} inference of \textit{large} MoE \tfm model on an edge device. 
It supports general MoE \tfm models for interactive tasks such as text generation and summarization. 
\sys mainly optimizes for transformer decoder, since it dominates the end-to-end inference time of MoE models due to its autoregressive nature (e.g., up to 93\% according to our experiments).

\sys incarnates as a runtime library linked to user apps. 
Along with \sys, MoE LLMs with \textit{experts compressed into different bitwidths} are also installed on an edge device.
It is configured by two key parameters. 
First, a memory budget $M$, specified either by users or the OS. 
The budget ranges from 1.5GB--3GB, 
which is one to two orders of magnitude smaller than the existing MoE LLMs. 
The flexible constraint accommodates varying device memory and adapts to system memory pressure.   
Second, a tolerable accuracy loss $P$ is chosen by the user. 
Based on the desired accuracy loss $P$, \sys tunes the individual bitwidths for the experts for constructing the model to be executed at run time.
Note this is a soft goal because existing MoE LLMs are unable to provide accuracy guarantees. 

Upon user invocation, \sys first selects the model satisfiable to accuracy loss $P$ and instantiates an expert preload/compute pipeline for reducing inference latency: 
it sequentially loads all non-expert weights by layers; 
depending on prior experts activated, it opportunistically loads the experts for the \textit{next} layers, overlapped with the computation of \textit{current} layers.
As a result of the inference, \sys generates a set of predicted tokens (e.g. in text generation task) or a summary (e.g. in summarization task).
\sys does not choose the bitwidths of experts at run-time, as it is difficult to predict the subsequent expert's impact on the overall accuracy, making it impractical to select the globally optimal bitwidth distribution.

During execution, \sys maintains two memory buffers:
1) an expert buffer used for managing and caching expert weights. 
It resides in memory along with \sys for supporting multiple rounds of inferences.    
2) a working buffer that holds all intermediate results. 
It is only temporary and can be thrown away right after each inference finishes.

\begin{figure}[t]
	\centering					
	\includegraphics[width=0.45\textwidth]{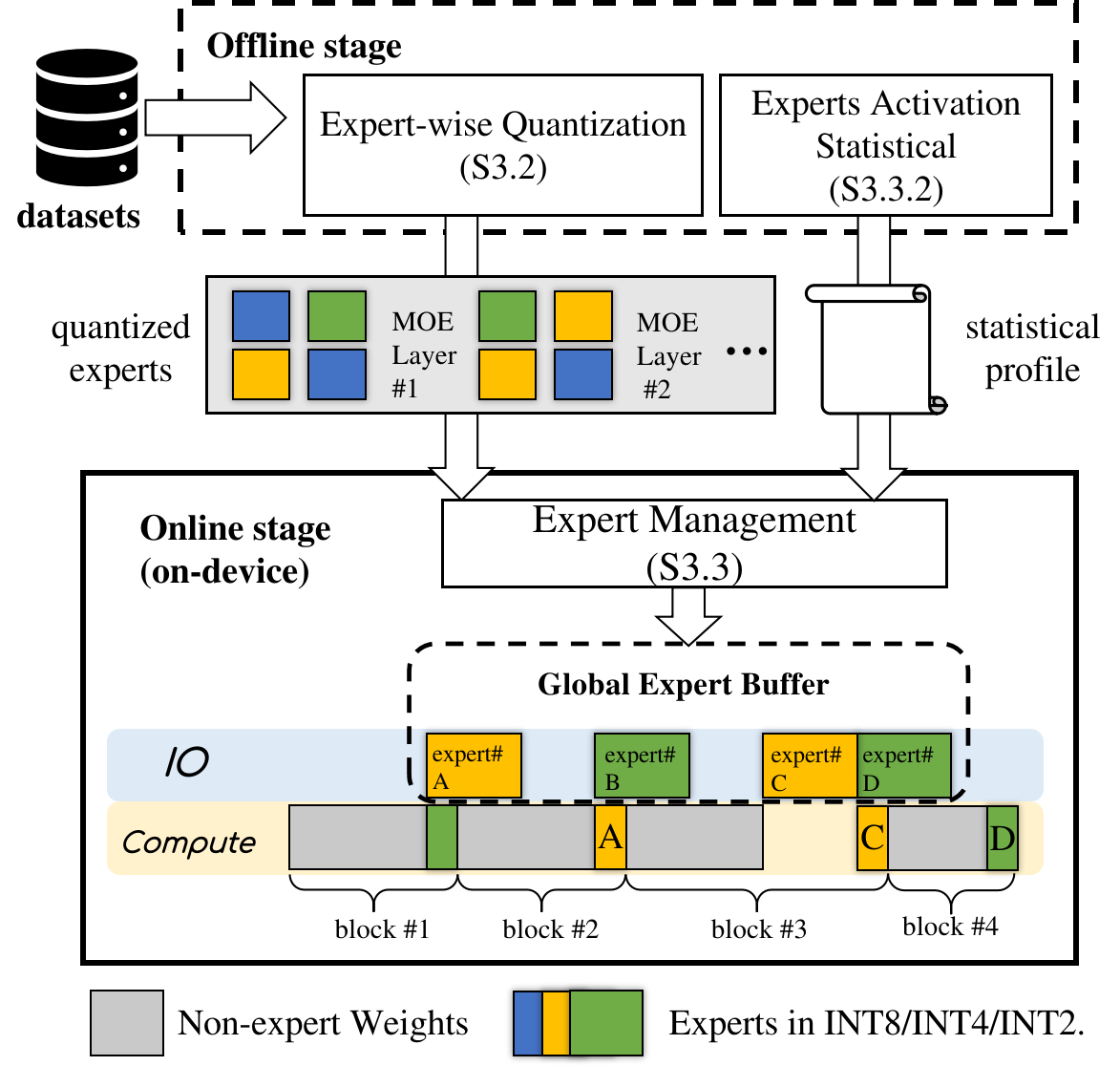}
	\caption{
 System architecture of \sys and workflow.
}
	\label{fig:workflow}
 \vspace{-10pt}
\end{figure}

\paragraph{The operation}
To use, \sys works in two main stages as shown in Figure~\ref{fig:workflow}.
\begin{myenumerate}

	\item 
	\textit{Offline expert quantization ($\S$\ref{sec:design:quantization}).}
	With an accuracy loss specified by the user (e.g. 5\%) on a given task, \sys first preprocesses a pre-trained model offline: it profiles the expert importance (the sensitivity to model accuracy) and then quantizes the experts to different bitwidths based on their assessed importance. 
	The resultant model comprises a set of experts with different bitwidths, even for those in the same transformer layer. 

	\item 
	\textit{Online expert management ($\S$\ref{sec:design:cache}).}
	At run time, \sys instantiates a preload/compute pipeline and dynamically manages the experts between device memory and disk via an expert buffer.
        By leveraging the statistical profile of expert activation, \sys pre-determines which experts to fetch from disk \textit{prior to} their router function and which to evict when the buffer is full.
        
 

\end{myenumerate}


\paragraph{Applicability}
The \sys framework is generic and applicable to both decoder-only and encoder-only transformer architecture. 
It is compatible with both dynamic (e.g. Switch Transformers~\cite{fedus2022switch}, GLaM~\cite{du2022glam}) and static routing (e.g. Hash layer~\cite{roller2021hash}, expert choice MoE~\cite{zhou2022mixture}) MoE layers.
Notably, in static routing layers, expert activation only depends on the original input tokens but not their hidden states.
For such layers, \sys simply preloads experts as instructed by input tokens in the pipeline without prediction.

The optimization of \sys is based on the sparsity of expert activation in the inference decoding stage of the MoE model, so it is suitable for accelerating the decoding stage of a single request. 
This is also the application scenario of the MoE \tfm model on edge devices such as mobile phones, such as using the MoE large language model for dialogue. 
However, for scenarios where the number of tokens in a single inference is greater than 1, which includes two scenarios: the prefiling stage and multiple input requests, \sys is not applicable.  
\sys can be used in combination with other efficiency optimization methods. For models with a large number of parameters and edge devices with poor computing performance, quantization can be adopted to optimize memory and computing speed.







\subsection{Expert-wise Quantization}\label{sec:design:quantization}

\begin{figure}[t]
	\centering
	\subfloat[ST-b-8, Xsum]{
		\includegraphics[width=0.23\textwidth]{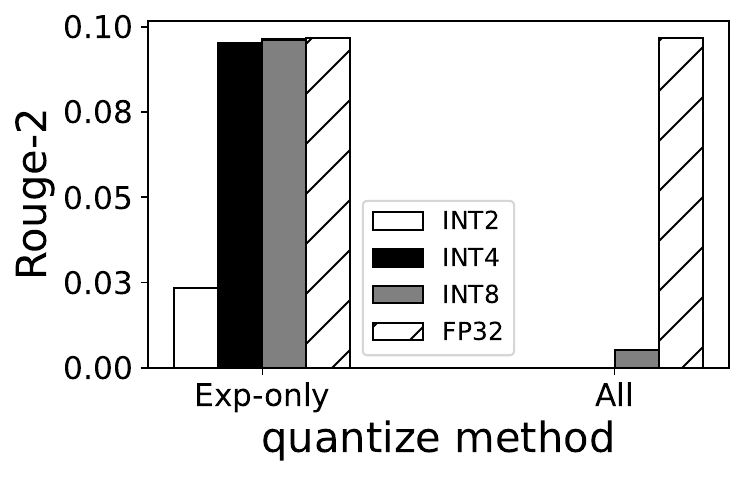}
	}
	\subfloat[ST-b-8, SAMsum]{
		\includegraphics[width=0.23\textwidth]{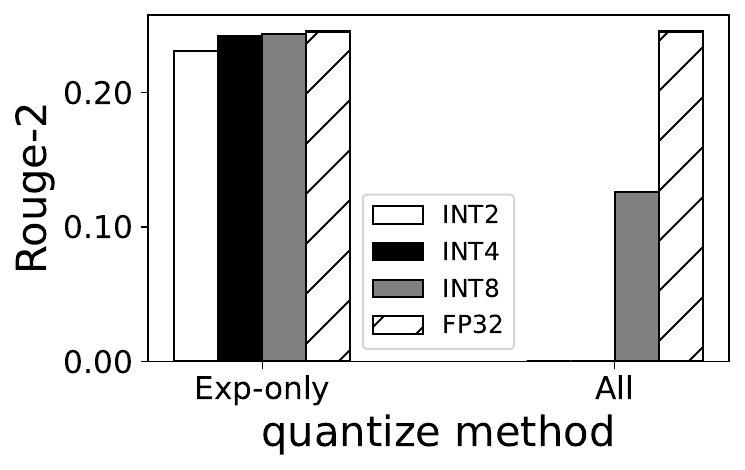}
	}
	\caption{
		The accuracy (Rouge-2) of quantizing experts weights/all weights into INT2/4/8.
	}
	\label{fig:motivation-quantizeMatric}
\end{figure}

To fit the expert weights under a set memory budget $M$ and to balance the compute/IO asymmetry, we opt for dictionary-based quantization~\cite{han2015deep}, which works with \textit{unmodified} pre-trained models;
we do not use quantization-aware training (QAT)~\cite{liu2023llm} to avoid retraining LLMs, which is tedious and expensive. 
While the quantization techniques are well-known, \sys is the first to apply them to \textit{individual} experts and to exploit accuracy vs bitwidths tradeoffs.



\begin{figure}[t]
	\centering
	\subfloat[Profiled expert importance]{
		\includegraphics[width=0.23\textwidth]{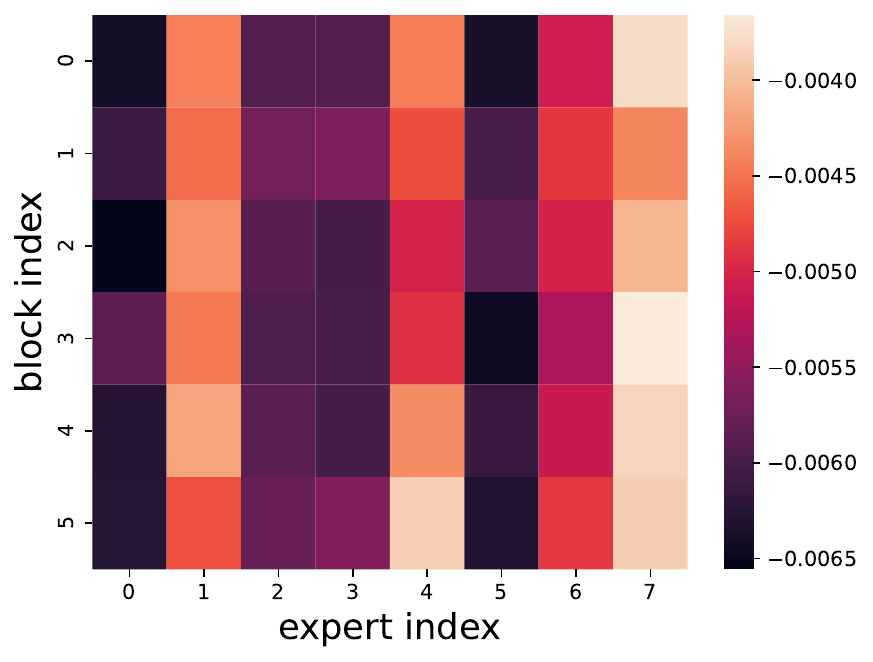}%
	}
	\subfloat[Measured decompression time]{
		\includegraphics[width=0.23\textwidth]{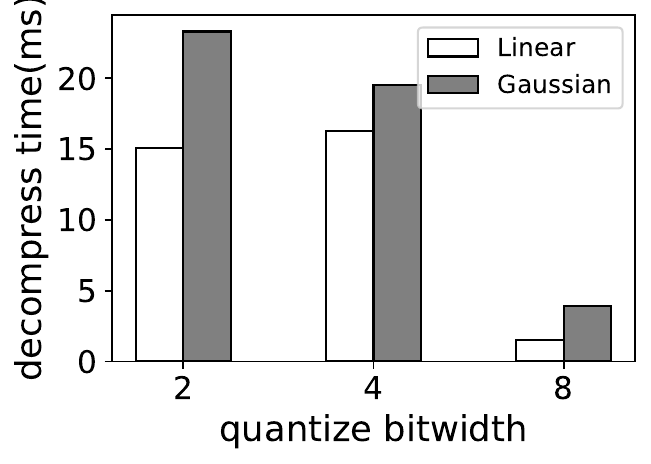}%
	}
	\caption{
		(a):The heatmap of accuracy loss.
        Each element in the heatmap represents the accuracy loss of the model when the expert parameter is quantified as INT2 and other expert parameters are quantified as INT4. For example, the first row and first column represent the accuracy loss when the first expert parameter of the first MoE layer of the model is quantized to INT2, while the other expert parameters are quantized to INT4.
		Model: ST-base-8. 
		Dataset: SAMsum.
		Accuracy: Rouge-2.
		(b):The decompression time of channel-wise linear quantization and gaussian outlier-aware quantization.
        The decompression time is measured on Jetson TX2.
	}
	\label{fig:design-quantime}
\end{figure}

\paragraph{Choosing the algorithm.}
We surveyed a wide range of quantization techniques (e.g. Gaussian outlier-aware quantization~\cite{zadeh2020gobo} and log-based quantization~\cite{aji2020compressing}) and have chosen channel-wise linear quantization~\cite{han2015deep} for its good accuracy and fast decompression time.
As shown in Figure~\ref{fig:motivation-quantizeMatric}, quantizing all experts weights to 4-bit integers (INT4) incurs only 1.30\%--1.44\% accuracy degradation;
on dataset SAMsum, the experts can be further quantized to 2-bit integers with only 5.82\% loss, acceptable to our use.
As shown in Figure~\ref{fig:design-quantime}(b), compared with other quantization techniques (e.g. Gaussian outlier-aware quantization~\cite{zadeh2020gobo}), channel-wise linear quantization is 1.1$\times$--2.5$\times$ faster, attributed to its simplified decompression process as a straightforward linear mapping.

Channel-wise linear quantization uniformly maps quantized integer values to the original float values using scaling factors. 
The scaling factor for each channel is determined by the maximum absolute value within that channel and the range expressible by the bitwidth.
This method requires an additional channel of float16 type scale to be reserved.
We have tested the same number of parameters on Jetson TX2. The dequantization time of channel-wise quantization is 2.7\% of the pure IO loading time, which is almost negligible. 



Quantized weights are not meant to be used as-is, which is different from QAT~\cite{liu2023llm}. 
Before use, we must decompress them,  which is a mirror process of compression. 




\paragraph{Profiling expert importance.}
For experts, we quantize them into different bitwidths, e.g. INT2/4/8.
The rationale is experts show different importance to model accuracy;
we want the most important experts to have high-bitwidth, hence contributing to the model accuracy more positively. 
\sys regards an expert as more important if it leads to the most accuracy loss when being executed in lower bitwidths.

To do so, \sys{} enumerates all experts, quantizes each to INT2, and profiles the resultant accuracy loss on different validation sets (e.g. ST-base-8). 
The results are shown as a heatmap in Figure~\ref{fig:design-quantime}(a).
For instance, quantizing the $1^{st}$ expert to INT2 at $1^{st}$ transformer block degrades the accuracy by 0.44\%, while quantizing $2^{nd}$ expert to the same precision causes 0.59\% degradation.
Therefore, the $2^{nd}$ expert is more sensitive to quantization and more important to model accuracy.
Such an observation is also backed up by prior literature~\cite{cui2023optimizing} that finds that different experts tend to handle different tasks with various difficulty levels.

As a result, \sys obtains the list of expert importance, which is sorted by the model accuracy when a corresponding expert is quantized into INT2.
The list will be used for constructing the runtime model, which we will shortly describe.



\paragraph{Selecting expert-wise bitwidths.}
Based on the user-tolerable accuracy loss, \sys judiciously selects the bitwidths of individual experts offline as follows.
\begin{myitemize}
	\item 
	First, \sys decides a bitwidth bound for \textit{all} experts of the model, which serves as a baseline for \sys to further tune. 
	To do so, \sys enumerates through the available bitwidths (e.g. INT2/4/8 and FP32) for all experts and measures the model accuracy, respectively.
	\sys then sets the lower and upper bound of bitwidths to those whose accuracies closely approximate the tolerable accuracy loss. 
	
	\item 
	Second, \sys tunes individually the bitwidth of experts based on the lower and upper bound of bitwidths. 
	It starts with the top-$K$ experts from the list obtained earlier.
	As they are less important, \sys quantizes them into lower bitwidth (i.e. INT2) while keeping the rest higher bitwidth (e.g. INT4).
	\sys then measures the accuracy of the resultant model. 
	If its accuracy loss is still lower than the desired goal, which means the model can afford more lower-bitwidth experts, \sys follows the list to gradually increase the parameters $K$ until the accuracy loss reaches the goal. 
	Otherwise, \sys decreases the parameters $K$ for reducing the accuracy loss, i.e. by lifting more experts to higher bitwidths.

\end{myitemize}

Through the above process, \sys obtains a model with mixed-precision experts, achieving a balance between accuracy and storage.
\mwx{Notably, \sys's choice of simplistic quantization algorithm implies that it is a training/finetuning-free approach, which is often regarded as a commendable advantage to LLM systems as it requires no training data.}


\paragraph{Non-expert weights quantization decision.}
Whether to quantize non-experts’ weights exhibits a sophisticated tradeoff:
it reduces memory usage, therefore leaving a larger cache room for experts;
meanwhile, it compromises model accuracy, therefore reducing the quantization space for experts given a tolerable accuracy loss.
\sys simply iterates over a few common candidates (INT4/INT8/FP16/FP32) using GPTQ algorithm~\cite{frantar2022gptq} to find out the one that exhibits the best performance.
Empirically, we find that non-experts' weights tend to be preserved at high fidelity (e.g., FP16) as they are activated during each inference.

\paragraph{\mwx{Complexity analysis.}}
\mwx{
The offline phase of \sys is dictated by the preset accuracy loss, as it requires iteratively enumerating various bitwidth configurations and dynamically adjusting expert assignments until the model’s accuracy drop meets the threshold. 
Meanwhile, the online phase merely applies the tuned configuration for parameter selection, incurring a much lower computational overhead and thereby satisfying real-time inference requirements.
}

\subsection{In-memory Expert Management}\label{sec:design:cache}

%
%

\begin{figure}[t]
	\centering
	\begin{minipage}[b]{0.23\textwidth}
		\includegraphics[width=\textwidth]{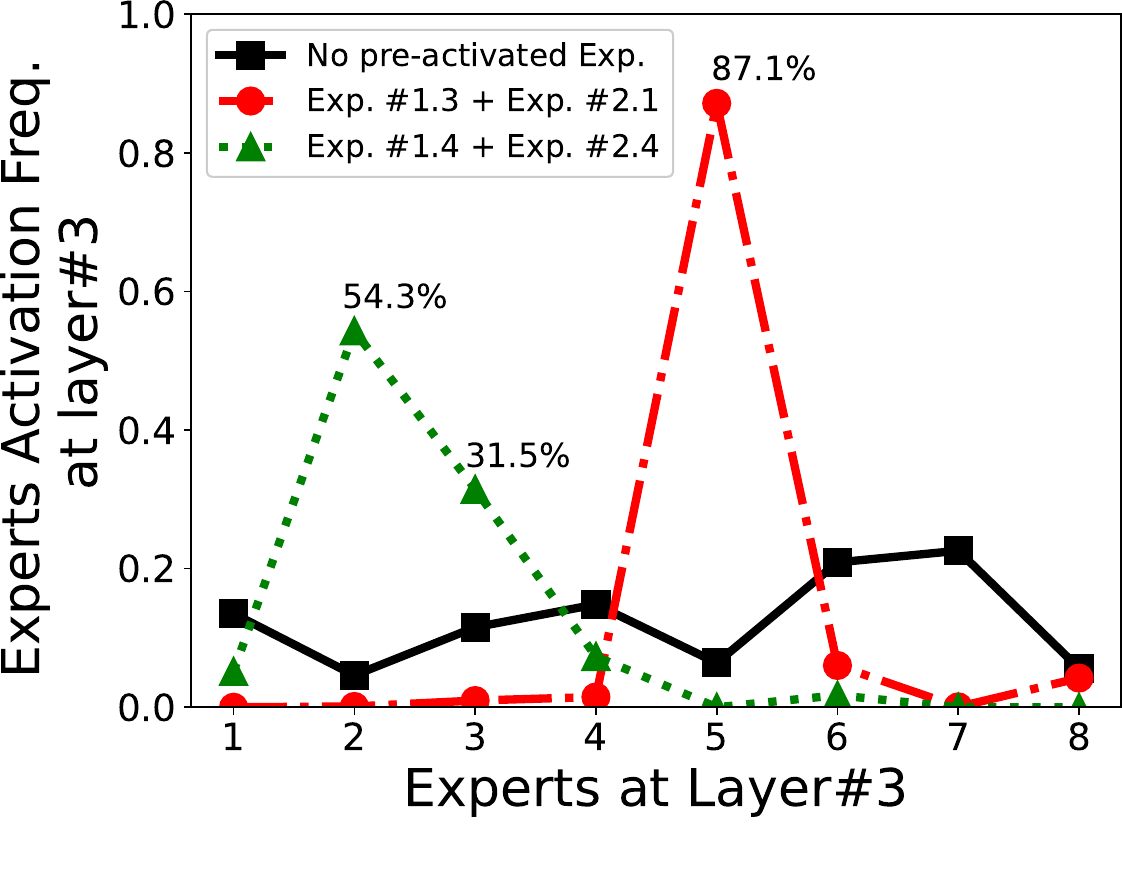}
	\end{minipage}
	~
	\begin{minipage}[b]{0.23\textwidth}
		\includegraphics[width=\textwidth]{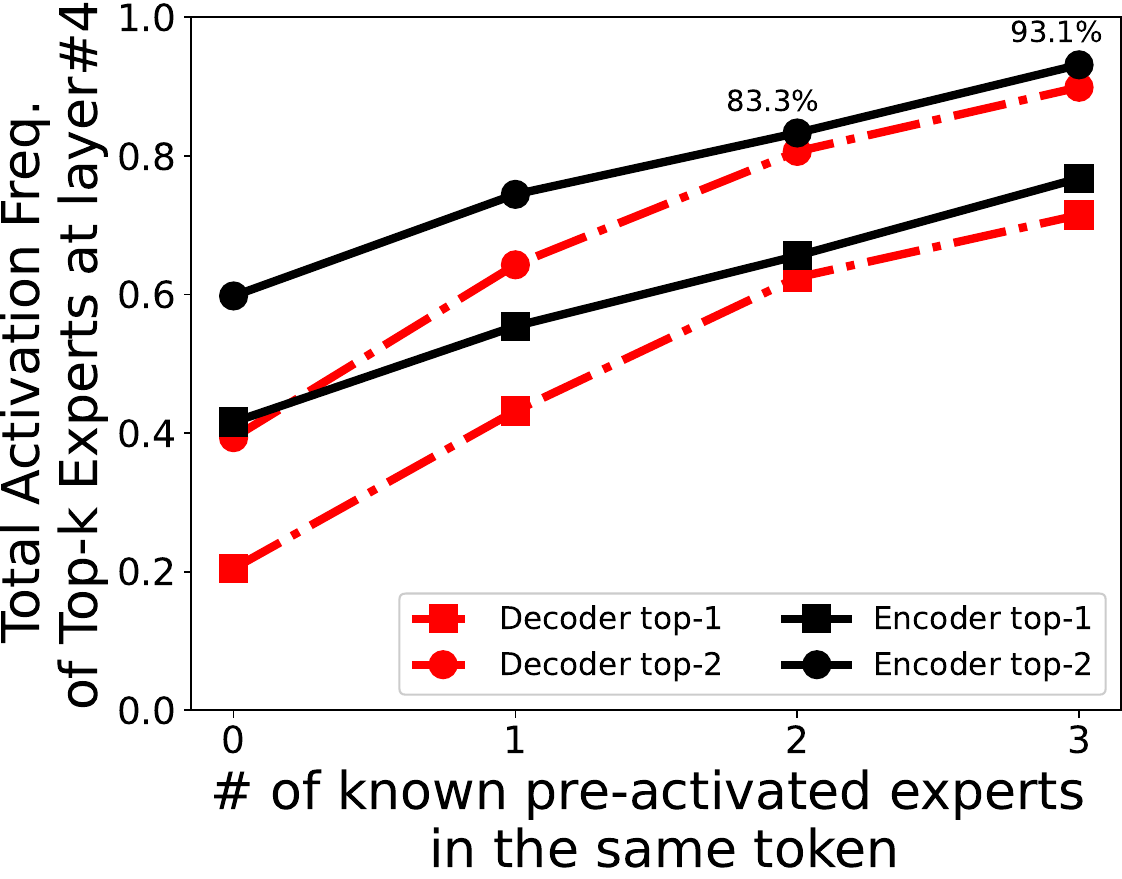}
	\end{minipage}
	\caption{
		Measurement results that demonstrate the expert activation correlation.
		(Left) The activation frequency of 8 experts at the 3rd decoder layer with and without knowing the activated experts at its first 2 layers. ``Exp \#X.Y'' indicates No.Y expert at X\textit{th} layer is activated.
		(Right) The accumulative activation frequency of the top-k experts by knowing different numbers of pre-activated experts at earlier layers in the same token.
		Model: ST-base-8; Dataset: SAMsum.
	}
	\label{fig:motivation-topk}
\end{figure}

%
%

\subsubsection{Preloading and pipeline}\label{sec:design:mm:act}
To overlap the expert loading with weight computation, we must \textit{predict} the expert activation beforehand, instead of passively waiting for the router output of previous MoE layers.


\textbf{Estimating expert activation a priori.} 
How to predict expert activation? 
We exploit a key observation that the expert activations of sequential layers are statistically correlated.
That is to say, \textit{given the prior knowledge of expert activations of $0..n-1$ layers}, we can estimate the activation probability of each expert at $n-th$ layer with good confidence, formulated as $P({E}_{n} = i | E_0, E_1, ..., E_{n-1})$ where $i$ is the index of the expert and $n$ is the layer index.
To demonstrate such a correlation, we analyze the expert activations running the ST-base-8 model on the SAMSum dataset.
As shown in Figure~\ref{fig:motivation-topk} (left), with two previous layers' activations observed, at layer $3$ there is a high probability (87.1\%) that No.5 expert will be activated, i.e., $P({E}_{3} = 5 | {E}_{1} = 3, {E}_{2} = 1) = 87.1\%$.
Figure~\ref{fig:motivation-topk}(right) further confirms this observation by statistically summarizing across different activation paths.

\textbf{Opportunistic preloading.}
\sys exploits the previous observation for opportunistically preloading the expert weights and executing the pipeline as follows. 
In the offline phase, based on the previous observation \sys executes the model on multiple datasets to build the statistical profile of expert activations. 
To this end, \sys generates a dictionary, 
wherein the \textit{key} denotes the activation status of experts from two previous consecutive MoE layers,
and the \textit{value} represents the probabilities of individual experts being activated in the subsequent MoE layer. 
The statistical profile is then stored for utilization in online inference.

In the online phase, before each MoE layer routing, \sys employs the activation statuses of experts in the previous layers as the \textit{key} for querying the statistical profile. 
Then, it sequentially preloads the experts to experts buffer (if not present) prioritized with their estimated activation probability.
The preloading stops until the router is finished and the real activated expert is thereby known. 
In practice, \sys can preload 1--3 experts in each layer for the pipeline, depending on the compute-I/O speedup gap.

\begin{figure}[t]
	\centering					
	\includegraphics[width=0.45\textwidth]{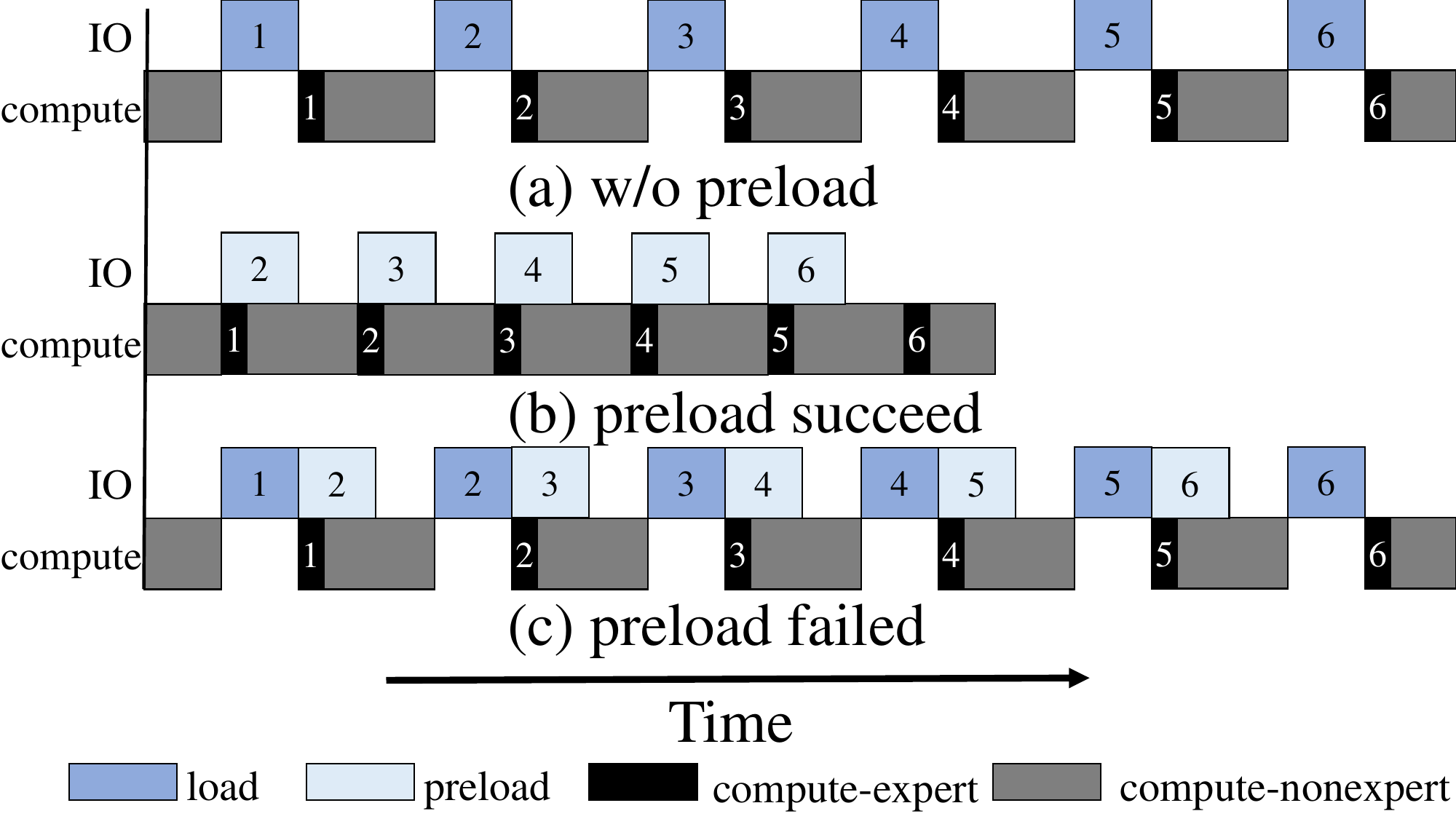}
	\caption{The pipeline scheduling for 1-token inference. 
	(a): Do not preload the next expert selection; 
	(b): Predict all expert selections successful; 
	(c): Predict all expert selection failed, which is the worst-case scenario.
	}
	\label{fig:design-preload}
\end{figure}
\textbf{The pipeline scheduling.}
\sys instantiates a preload/compute parallelism:
it executes the computations within the current transformer block and the preloading of the subsequent transformer block in parallel, based on the prediction made by the statistical profile.
Figure~\ref{fig:design-preload} elucidates the pipeline scheduling for situations where prediction is both successful/failed.
When \sys accurately forecasts the activation of the next MoE layer's expert which is a common case, it significantly reduces the end-to-end inference delay, hiding the loading time under computation. 
As a worst-case scenario (which we have never observed), when all predictions fail, the inference time is only as long as loading experts on demand. 

\subsubsection{Cache eviction policy}\label{sec:design-mm:cache}

\sys maintains a cache for expert weights in memory, sized by the memory budget $M$.
Once an in-cache expert is activated, the weights are directly fetched from the buffer, i.e., a cache hit.
Otherwise, \sys needs to fetch the expert from disk and evict an old expert when the cache is full.
The cache eviction policy -- determining which expert to evict, is crucial to \sys performance since wrongly evicting an expert that will be used shortly causes significant I/O overhead.

Classic cache policies like FIFO/LRU/LFU are designed for operating systems, mainly based on the data access history and frequency.
\sys leverages the expert activation frequency as well, yet incorporates another unique opportunity: since LLMs are built with sequentially stacked transformer layers, each expert's activation timing co-relates with its position (i.e., layer index).
Specifically, if an expert resides in a layer that is going to be executed soon, it shall be endowed with a higher score for not being evicted.

Based on this heuristic, \sys's eviction policy considers both the frequency of expert usage and the MoE layer index. 
The key idea is to give priority to the eviction of experts stored in the buffer with lower usage frequency and those whose layers are farther away from the current block.
We formulate the eviction policy as follows. 
For the $j$-th expert at $i$-th MoE layer, we define its eviction score as $L_{i,j}$: 
$$ L_{i, j} =-\frac{f_{i, j} }{{(S-i+I) \mod S}}$$
where $I$ is the index of current MoE layer, $f_{i, j}$ is the frequency of the $j$-th expert activated in the $i$-th MoE layer.
And $S$ indicates the size of MoE layers in this model's decoder.
Therefore, the higher the score $L$ is, the more likely the expert will be evicted. 
For the experts within the encoder, we set the frequency $f_{i, j}$ to 0. 
The reason is these experts are loaded only once after re-batching, so they should be prioritized for eviction.
When initializing the expert buffer, we load encoder expert weights sequentially.
For encoder-only models, the expert buffer is initialized with the experts having the highest frequency of usage.

\section{Evaluation} 
\label{sec:eval}

\subsection{Implementation and Methodology}\label{sec:eval:setup}

\begin{table}[t]
	\scriptsize
	\begin{tabular}{l|cccccc|c}
		\hline
		\multicolumn{1}{l|}{\textbf{Model}}& \multicolumn{1}{l}{\textbf{Type}} & \multicolumn{1}{l}{\textbf{EnM/En}} & \multicolumn{1}{l}{\textbf{DeM/De}}& \multicolumn{1}{l}{\textbf{Exp.}}  & \multicolumn{1}{l}{\textbf{K}} & \multicolumn{1}{l|}{\textbf{Params.}} & \multicolumn{1}{l}{\textbf{olT(min)}} 			\\ \hline
		ST-b-8  & en-de   & 6/12  & 6/12  & 8  & 1 & 0.4B 	& 7.7 	\\ \hline
		ST-b-16	& en-de   & 6/12  & 6/12  & 16 & 1 & 0.9B	& 9.4		\\ \hline
		ST-b-32	& en-de   & 6/12  & 6/12  & 32 & 1 &1.8B 	& 13.2		\\ \hline
		ST-b-64	& en-de   & 6/12  & 6/12  & 32 & 1 & 3.5B	& 19.5		\\ \hline
		ST-b-128& en-de   & 6/12  & 6/12  & 128 & 1 & 7.1B	& 23.3		\\ \hline
		ST-l-128& en-de   & 12/24  & 12/24  & 128 & 1 & 26B	& 28.7		\\ \hline
		GPTSAN	& de      & 0/0   & 9/9     & 16  & 2 &0.6B	& 8.2		\\ \hline
\end{tabular}
\caption{
	MoE models used in experiments.
	"ST-b-X": Switch Transformers base model with X experts per MoE layer.
	"ST-l-X": Switch Transformers large model with X experts per MoE layer.
	"En": number of encoders;
	"De": number of decoders;
	"EnM": number of MoE layers in encoders;
	"DeM": number of MoE layers in decoders.
	"Exp.": number of experts in each MoE layer;
	"K": top-k experts in each MoE layer;
	"Params.": number of parameters;
	"olT.": average off-line time (min).
	}
\vspace{-10pt}
\label{tab:exp-models}
\end{table}

\noindent \textbf{\sys prototype}
We've fully implemented a prototype of \sys with 1K Python LoC atop transformers. 
We used Pytorch as the transformers' backend and CUDA backend for its more generic support for different platforms. 
We will use a separate thread independent of the computing (main thread) for preloading. To prevent this IO-intensive thread from seizing the computing resources of the computing thread, it is bound to the small cores of the CPU. Since the GPU on Jetson TX2 performs dequantization more slowly than the CPU, we handle it on the CPU instead. We defined a map to represent the expert's buffer. 
At the same time, for smartphone execution, we implemented \sys in the self-developed C++-based large language model inference engine mllm~\cite{yi2023mllm}.
Note that the techniques of \sys are also compatible with other DL libraries.

\noindent \textbf{Models}
We use 7 popular MoE-based sparse LLMs as summarized in Table~\ref{tab:exp-models} to test the performance of \sys.
Most of these models are based on Switch Transformers~\cite{fedus2022switch} architecture in encoder-decoder structure with top-1 routing, i.e., only 1 expert is activated per layer.
Besides, GPTSAN~\cite{tanrei-gptsan-jp} has a decoder-only structure and works as a shifted Masked Language Model for Prefix Input tokens. 
It uses top-2 routing.
We obtain the pre-trained models from Hugging Face.~\cite{huggingfacehubmodels}.

\noindent \textbf{Datasets}
We evaluate \sys with three NLP downstream datasets:
(1) Xsum Dataset~\cite{huggingface-xsum}: Comprising a substantial collection of 226,711 news articles, each accompanied by a concise one-sentence summary.
(2) SAMsum Dataset~\cite{huggingface-samsum}: This dataset features approximately 16,000 conversation transcripts, reminiscent of messenger exchanges, along with corresponding summaries.
(3) Wikipedia-jp Dataset~\cite{huggingface-wikipedia-japanese}: This extensive dataset encompasses the entire corpus of Japanese Wikipedia articles as of August 8, 2022.
Datasets Xsum and SAMsum are specifically employed for the summarization task, where the objective is to generate a summary of the input content. We evaluated the performance of the Switch Transformers model on these datasets.
Conversely, the Wikipedia-jp dataset serves as the foundation for text generation tasks. We assessed the capabilities of GPTSAN in text generation tasks using this dataset.

\noindent \textbf{Metrics}
We mainly report the model accuracy, inference speed (per token and sequence), peak memory footprint, and model size of \sys and baselines.
To assess model accuracy, we use the Rouge-2 metric~\cite{lin2005recall} in our experiments. 
It comprises a collection of metrics designed for the evaluation of automatic summarization and text generation tasks. 
In the context of summarization, Rouge-2 quantifies similarity by comparing automatically generated abstracts with a reference set of abstracts, typically manually crafted. 

\noindent \textbf{Hardware}
We evaluate \sys on two prominent edge devices: the Jetson TX2 (GPU) and the Raspberry Pi 4B (CPU).
Both the Jetson TX2~\cite{tx2} and Raspberry Pi 4B~\cite{rpi4b} run Ubuntu 18.04 as their operating system. 
Since MoE LLMs are large, we need external storage to hold them.
For Raspberry Pi 4B, we use SDCards (SanDisk Ultra 128GB~\cite{sdcard});
for Jetson TX2, we use two types of hard drives, namely SSD (default) and HDD. 
The SSD model is SAMSUNG 860 EVO~\cite{samsungssd}, boasting a read/write speed of 550/520 MB/s. 
and HDD model is MOVE SPEED YSUTSJ-64G2S~\cite{movespeed} who provide a read/write speed of 50/20 MB/s.
The offline stage of \sys to generate a quantized MoE is performed on a GPU server equipped with 8x NVIDIA A40. 

\noindent \textbf{Baselines}
We compare \sys with four baselines. 
(1) \texttt{IO-FREE} assumes all model weights are held in memory so no swapping I/O is needed.
This is the most computationally efficient approach but is not scalable due to memory constraints.
(2) \texttt{IO-QFREE} quantizes the parameters of the model into INT8 using hannel-wise linear quantization and loads the parameters and executes in the way of IO-FREE. The quantization is achieved through Quanto~\cite{quanto-introduction}.
(3) \texttt{IO-EXP} treats memory as a cache for experts and dynamically loads them once activated, similar to \sys.
(4) \texttt{IO-QEXP} combines the above method with MoQE~\cite{kim2022mixture} to
quantize experts weights into INT4 and dynamically loading them during inference.
Alike \sys, the quantized weights need to be converted back to FP32 for fast inference on device processors.
(5) \texttt{STI} minimizes inference accuracy by model sharding and instantiates an IO/compute pipeline under a tight memory budget~\cite{guo2023sti}.
It does not differentiate the weights for experts and non-experts.
For a fair comparison, we adjust the size of the buffer for preload shards so that STI and \sys have the same memory footprint.

\noindent \textbf{Configurations}
If not otherwise specified, we set the expert buffer of \sys to 10$\times$ experts; the tolerable accuracy loss is 2\%.
Each experiment is conducted systematically with multiple repetitions, and the reported values are based on their respective averages.

\subsection{End-to-end Results}

\begin{figure*}[t]
	\centering					
	\includegraphics[width=0.95\textwidth]{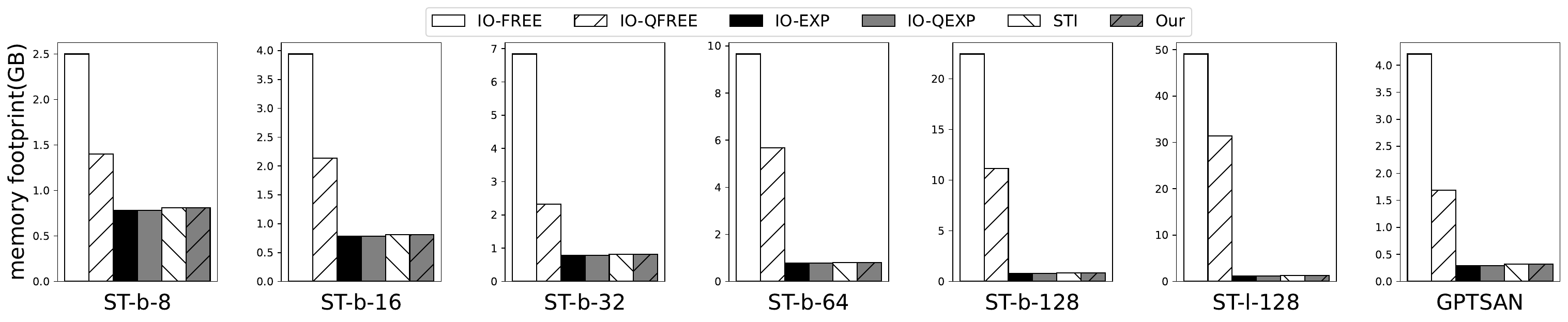}
	\vspace{-10pt}
	\caption{The memory footprint of \sys and baselines.}
	\label{fig:eval-mem}
\end{figure*}
\begin{figure*}[t]
	\centering
	\subfloat[Jetson TX2]{
		\includegraphics[width=0.95\textwidth]{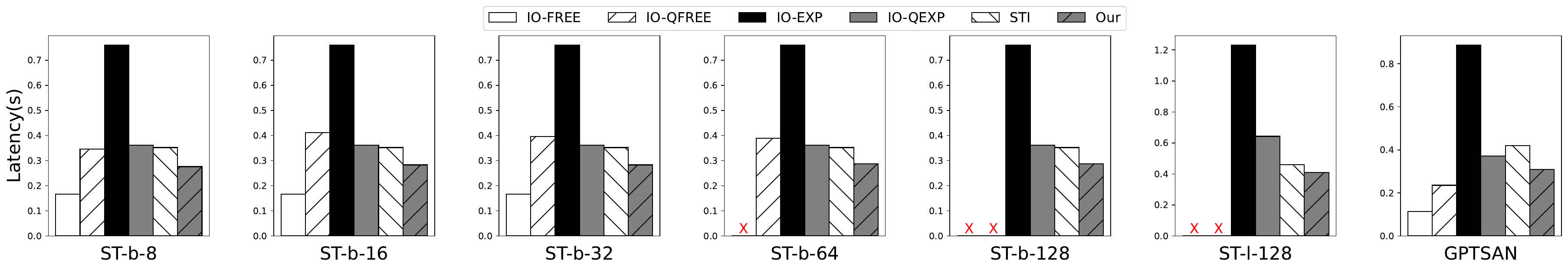}
	}
	\hfil
	\subfloat[Raspberry Pi 4B]{
		\includegraphics[width=0.95\textwidth]{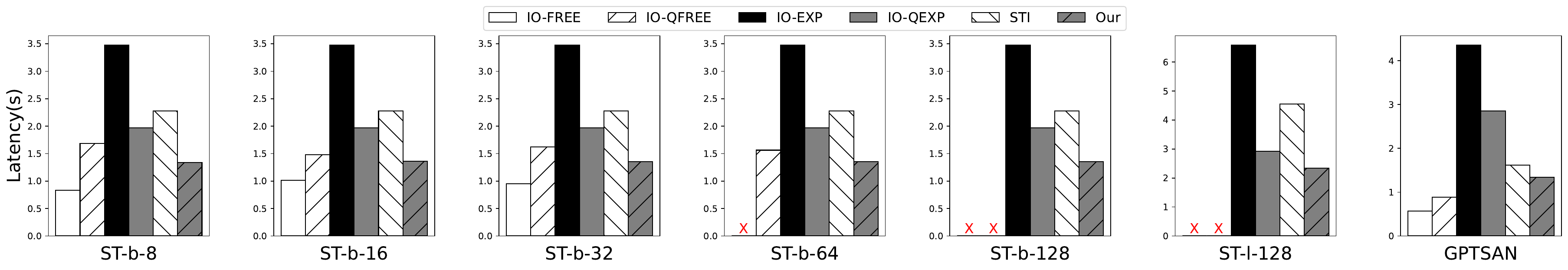}
	}
	\hfil
	\subfloat[Jetson TX2 - HDD]{
		\includegraphics[width=0.95\textwidth]{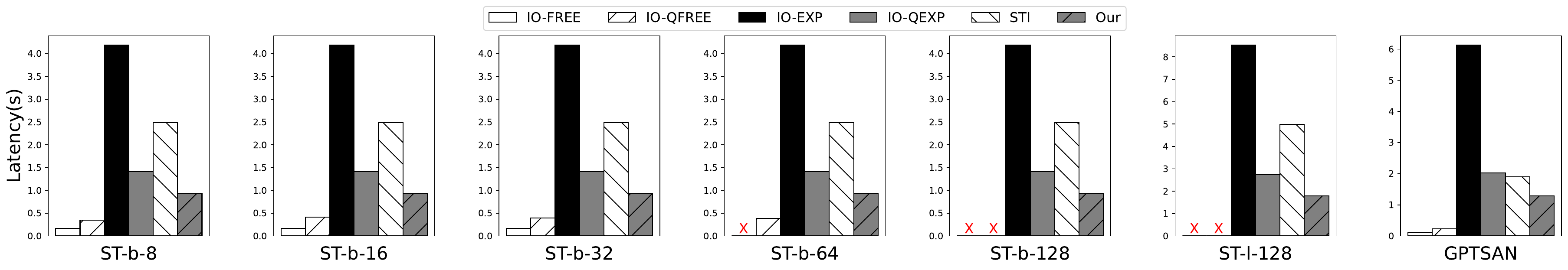}
	}
	\caption{
            Time Per Output Token (TPOT) of \sys and baselines in edge devices. 
		The baseline with a red "X" symbol above the bars could not be accommodated within the memory constraints of the target hardware. 
		The height of the bars represents the theoretically predicted values.
            "ST" is the abbreviation of Switch Transformers. "b" is the abbreviation of base, and "l" is the abbreviation of large. For example, "ST-b-8" represents Switch Transformers base-8.  
		}
	\label{fig:eval-time}
\end{figure*}
\begin{figure}[t]
	\centering					
	\subfloat[Jetson TX2]{
		\includegraphics[width=0.23\textwidth]{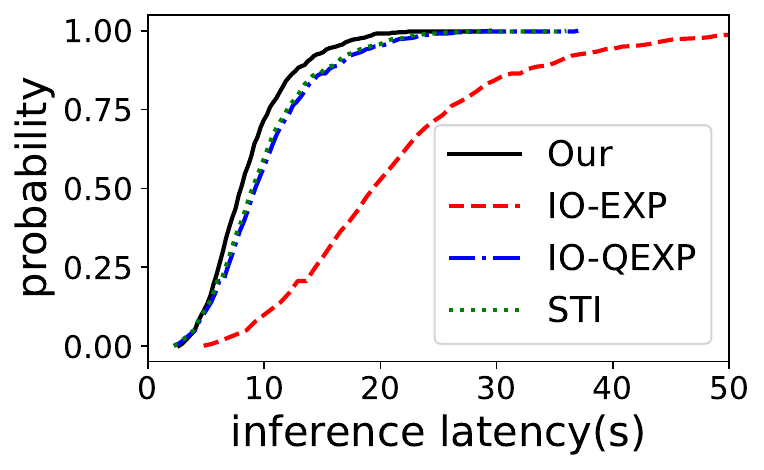}
	}
	\subfloat[Raspberry Pi 4B]{
		\includegraphics[width=0.23\textwidth]{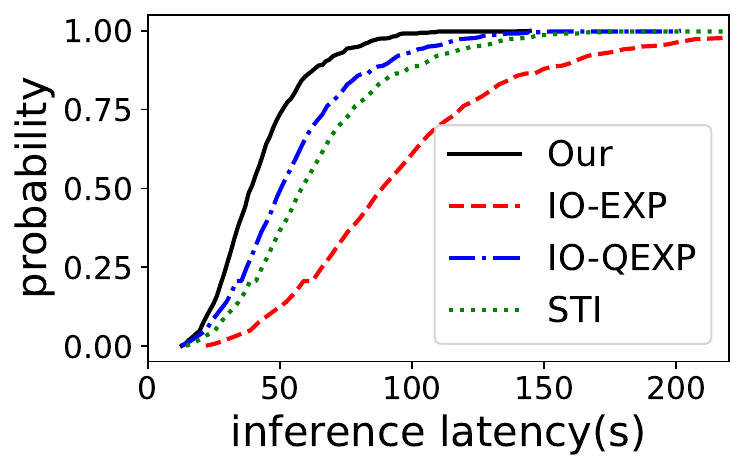}
	}

	\caption{
		The cumulative distribution function (CDF) of per-sample inference latency.
		Model:ST-base-8.
	}
	\label{fig:eval-timecdf}
\end{figure}

\textbf{Memory footprint.}
We conducted a memory footprint evaluation of \sys and the baselines on edge devices. 
The results are shown in Figure~\ref{fig:eval-mem}. 
\sys significantly outperforms the baselines across all models and platforms, achieving memory savings ranging from 2.1$\times$ to 38.6$\times$ compared to IO-FREE, and 0.8$\times$ to 24.3$\times$ compared to IO-QFREE.
This improvement can be attributed to \sys's efficient management of inactive expert weights and activations. 
Additionally, \sys dynamically loads and caches activated expert weights in memory to reduce memory footprint.

In contrast, \sys exhibits a memory footprint similar to that of IO-EXP and IO-QEXP. 
Since these two baselines do not require caching prior expert weights, \sys incurs a slightly higher memory footprint in comparison. 
For example, when the expert buffer is set to 10$\times$ the expert's memory footprint, ST-base models consume approximately 180MB more memory than IO-EXP and IO-QEXP. 
According to the settings outlined in $\S$\ref{sec:eval:setup}, the baseline STI shares the same memory footprint as \sys.

\textbf{Time Per Output Token.}
Figure~\ref{fig:eval-time} compares end-to-end Time Per Output Token(TPOT) between \sys and the baselines on edge devices. 
The weights decompression costs are contained in the latency.
The results highlight a significant performance improvement achieved by \sys across all models and platforms. 
In Jetson TX2, \sys demonstrates a speedup ranging from 2.64$\times$ to 3.03$\times$ compared to IO-EXP, and in Raspberry Pi 4B, the speedup ranges from 2.55$\times$ to 3.25$\times$. 
This notable performance gain can be attributed to several key factors. 
Firstly, \sys employs weight quantization for the experts, effectively reducing loading latency. 
Additionally, \sys adopts an efficient strategy for preloading expert weights, intelligently overlapping this preloading process with computation, thereby effectively masking most of the latency. 
Consequently, \sys achieves a commendable speedup of 1.19$\times$--2.12$\times$ compared to IO-QEXP and 1.22$\times$--2.77$\times$ inference speedup over STI.
\sys has an advantage over STI in TPOT. STI's technology is independent of the model structure and does not utilize the feature that only some expert parameters of the MoE model are activated during inference. So, all parameters are dynamically loaded during inference, which leads to a lot of time spent loading parameters of unactivated expert parameters.

However, a performance gap still exists between \sys and IO-FREE because \sys's preloading stage doesn't always predict which expert to activate for the next MoE layer. 
Some experts still need to be loaded dynamically.
On Jetson TX2 (GPU), IO-QFREE has higher latency than \sys, but on Raspberry Pi (CPU), IO-QFREE is faster. This may be because the Jetson TX2 GPU is outdated, and its int-to-float inverse quantization process is slow.


\textbf{Impact of IO speed}
Figure~\ref{fig:eval-time} also compares per-token inference latency between SSD and HDD on Jetson TX2. 
Notably, \sys achieves a higher acceleration rate on lower-cost HDDs compared to SSDs, especially when compared to baselines IO-EXP. 
For example, compared to IO-EXP, \sys achieves a speedup ranging from 4.49$\times$ to 4.76$\times$ on HDDs and from 2.63$\times$ to 3.01$\times$ on SSDs. 
This difference is due to the relatively slower read speeds of HDDs, resulting in longer expert weight loading times compared to SSDs. 
\sys demonstrates a more significant improvement in expert loading, leading to a more pronounced enhancement in per-token inference latency.

\textbf{Per-sample inference latency.}
We also evaluate the per-sample inference latency of \sys compared to the baselines on both Jetson TX2 and Raspberry Pi 4B. 
The cumulative distribution function (CDF) for the ST-base-8 model is depicted in Figure~\ref{fig:eval-timecdf}. 
The results show that \sys consistently outperforms the baselines. 
For instance, on the Raspberry Pi 4B, 50\% of the samples processed by \sys exhibit a latency of less than 46 seconds, whereas with IO-EXP, 50\% of the samples experience a latency of less than 106 seconds.
Same as Per-token inference latency, the performance gap still exists between \sys and IO-FREE.

\textbf{Offline profiling/quantization time.}
In the offline stage, \sys profiles and explores the per-expert quantization design space.
We utilized a dataset which contains 410 instance from Xsum and 409 instance from SAMsum, 
with a batch size set to 8.
As shown in Table~\ref{tab:exp-models} (last column),
the offline stage can be completed within half an hour, demonstrating an acceptable overhead offline. 

\begin{table}[h!]
    \centering
    \subfloat[ST-base-8]{
        \resizebox{0.24\textwidth}{!}{%
            \begin{tabular}{l|ccc}  
                \hline
                baseline      &Storage(GB)   &Loss(\%)  &Rouge-2   \\ \hline
                IO-FREE     &2.43   &0.00   &0.221 \\ \hline
                IO-EXP      &2.53   &0.00   &0.221  \\ \hline
                IO-QEXP     &0.85   &2.04   &0.215\\ \hline
                STI         &0.85   &20.0   &0.176\\ \hline
                Our         &0.81   &0.89   &0.219   \\ \hline
            \end{tabular} 
        }
    }
    \subfloat[ST-base-16]{
        \resizebox{0.24\textwidth}{!}{%
            \begin{tabular}{l|ccc}  
                \hline
                baseline      &Storage(GB)   &Loss(\%)   &Rouge-2    \\ \hline
                IO-FREE     &4.12   &0.00   &0.245 \\ \hline
                IO-EXP      &4.12   &0.00   &0.245 \\ \hline
                IO-QEXP     &1.03   &3.15   &0.237\\ \hline
                STI         &1.03   &25.1   &0.183\\ \hline
                Our         &0.95   &1.95   &0.240\\ \hline
            \end{tabular} 
        }
    }
    \caption{The storage and accuracy (Rouge-2) loss of \sys and baselines.
    "Loss" indicates accuracy loss.}
    \label{tab:eval-storageacc}
\end{table}

\textbf{Storage.}  
We also compared the storage requirements of \sys and the baselines while measuring their accuracy loss. 
Table~\ref{tab:eval-storageacc} presents the experimental results for ST-base-8 and ST-base-16. 
Notably, \sys significantly outperforms the baselines in terms of storage, achieving a 3.03$\times$ improvement over IO-FREE and a 1.11$\times$ improvement over IO-QEXP for ST-base-8. 
This superiority stems from \sys's utilization of a mixed-precision quantization method for expert weights.

\textbf{Accuracy.}  
Furthermore, the accuracy loss aligns with our expectations. 
\sys exhibits an accuracy loss that closely approximates the tolerable 2\% threshold. 
IO-FREE, IO-EXP, and IO-QEXP models similarly show accuracy losses consistent with their respective bitwidth configurations. 
Unlike the other baselines, STI quantizes both non-expert weights and expert weights, leading to a significant accuracy loss.

In the Appendix A, we show two comparative examples of the results with or without using \sys. The model used is Switch Transformers base-8, and the task is "summary". Example 1 is a case with a poorer effect and it can be seen that the output of \sys is quite different from the original version. The quantization of the expert parameter causes all the accuracy errors. \sys will reload experts that were not predicted correctly, so there is no result error caused by loading the wrong expert

\begin{figure}[t]
	\centering
	\subfloat[Jetson TX2]{
		\includegraphics[width=0.23\textwidth]{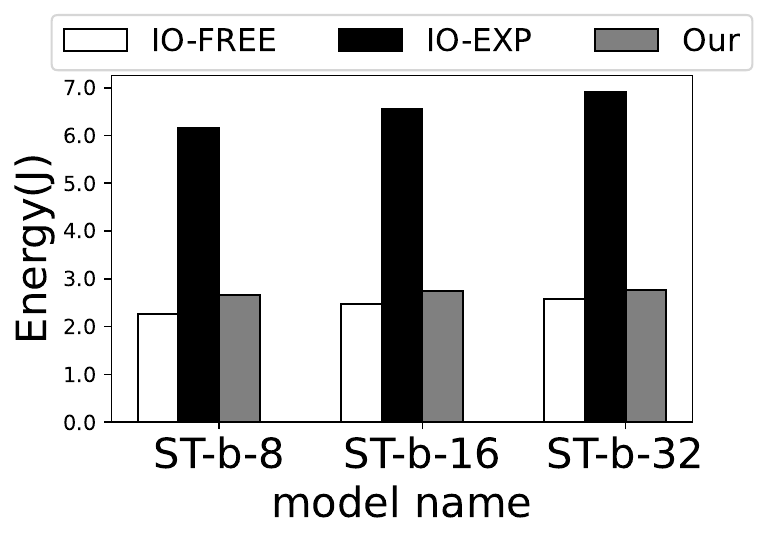}
	}
	\subfloat[Raspberry Pi 4B]{
		\includegraphics[width=0.23\textwidth]{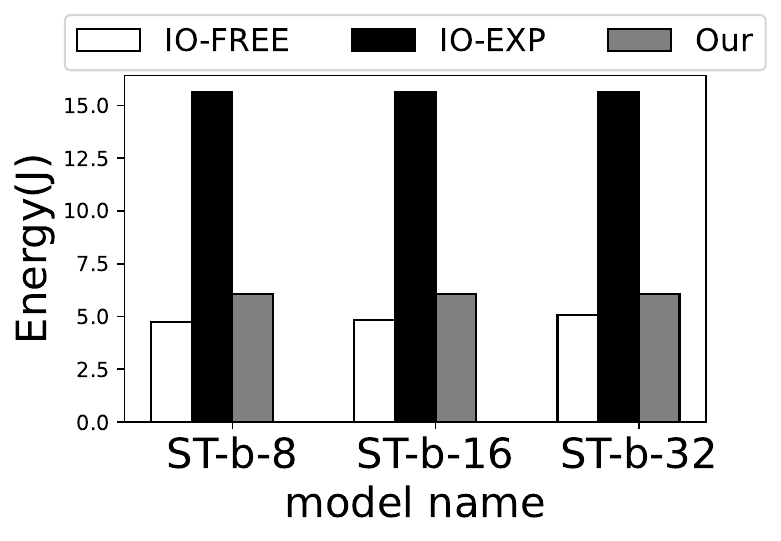}
	}
	\caption{
		The energy cost of per-token inference.
	}
	\label{fig:eval-energy}
\end{figure}
\textbf{Energy overhead.}
We measured the energy consumption of token inference on \sys across two edge devices and compared it with two baselines, IO-FREE and IO-EXP, as shown in Figure~\ref{fig:eval-energy}. 
Energy consumption is obtained by multiplying power by execution time. 
Compared to IO-FREE, \sys incurs, as small as 1.1\%, on average 17\% increase. The faster the storage is, the larger the model is, \sys's energy overhead is smaller. The reason is inference is compute-intensive, where CPU/GPUs consume major energy.

\begin{figure}[t]
	\centering					
	\subfloat[ST-base-8]{
		\includegraphics[width=0.23\textwidth]{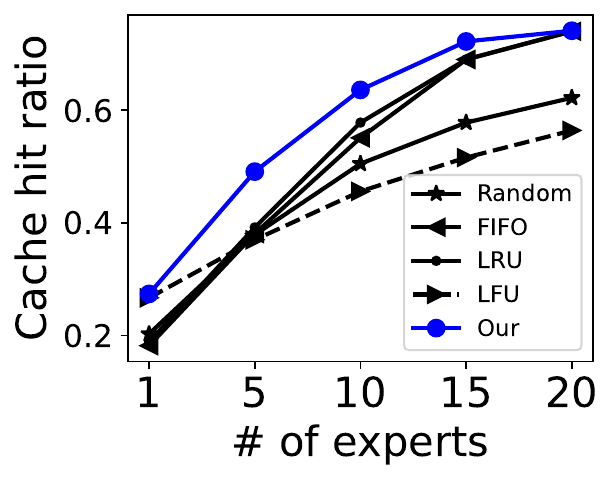}%
	}
	\subfloat[ST-base-16]{
		\includegraphics[width=0.23\textwidth]{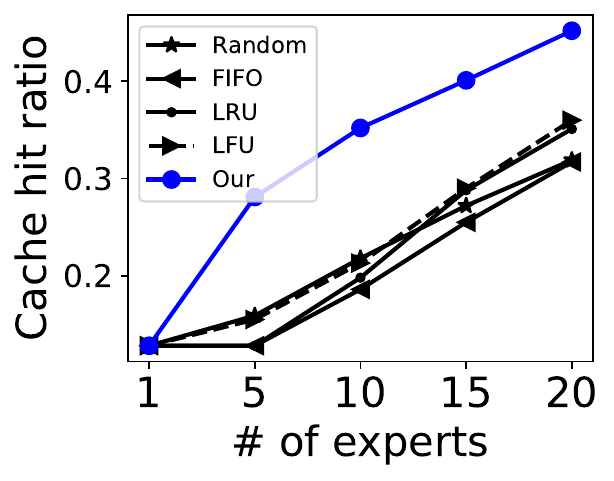}%
	}
	\caption{
		Cache hit ratio for 5 cache eviction policy. 
	"\# of experts" indicates the number of experts can be saved in expert buffer. }
	\label{fig:design-hitrate}
\end{figure}
\textbf{Cache eviction policy.} 
We perform a comparative analysis of cache hit ratios between our novel cache eviction policy and other policies with varying expert buffer sizes. 
To mitigate the influence of preloading on the hit ratio, we disable the preloading functionality in these experiments. 
The results are shown in Figure~\ref{fig:design-hitrate}. 
Our novel eviction policy exhibits superior efficacy compared to several other policies.

\begin{figure}[t]
	\centering
	\subfloat[Latency]{
		\includegraphics[width=0.23\textwidth]{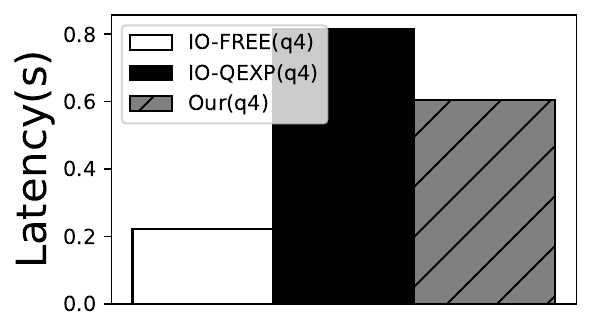}
	}
	\subfloat[Memory]{
		\includegraphics[width=0.23\textwidth]{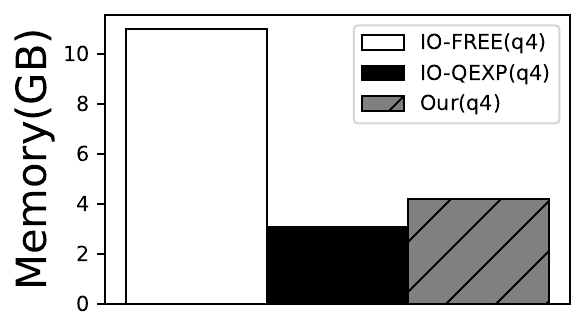}
	}
	\caption{
        The experiments of \sys for MiniCPM MoE 8x2B model and the baseline running on the Xiaomi 14 mobile phone. The usage of TPOT and memory is respectively presented.
	}
	\label{fig:eval-phone}
\end{figure}

\textbf{Performance on Android mobile phones.} 
We used the mllm~\cite{yi2023mllm}, a C++-based large language model inference engine to implement \sys, supporting the MiniCPM MoE 8x2B model~\cite{minicpmmoe8x2b}. 
The smartphone we used in our experiment was Xiaomi 14, which has 16GB of RAM and Snapdragon 8Gen3 SoC.
\sys is orthogonal to model quantization works such as GPTQ and AWQ and can be used in a superimposed manner. This is because \sys only affects experts' parameters. Non-expert parameters can also be quantized.  
Above all, when conducting experiments on the mobile phone and applying \sys, we performed group quantization of non-expert parameters in groups of 32 as INT4, and executed it using 4 threads on 4 large CPU cores. Baseline data compared to \sys include IO-FREE and IO-QEXP.

The experimental results are shown in Figure~\ref{fig:eval-phone}. It can be seen that on the mobile phone, compared to IO-FREE, 63\% of memory usage is saved, and compared with IO-QEXP, the speed increases by 25\%. 
The experimental effect of \sys on mobile phones is not as obvious as that on Jetson devices. The main reason is that the computing performance of the mobile phone chip is powerful and IO has become a bigger bottleneck.

\subsection{Sensitivity Analysis}

\begin{figure}[t]
	\centering
	\subfloat[Jetson TX2]{
		\includegraphics[width=0.45\textwidth]{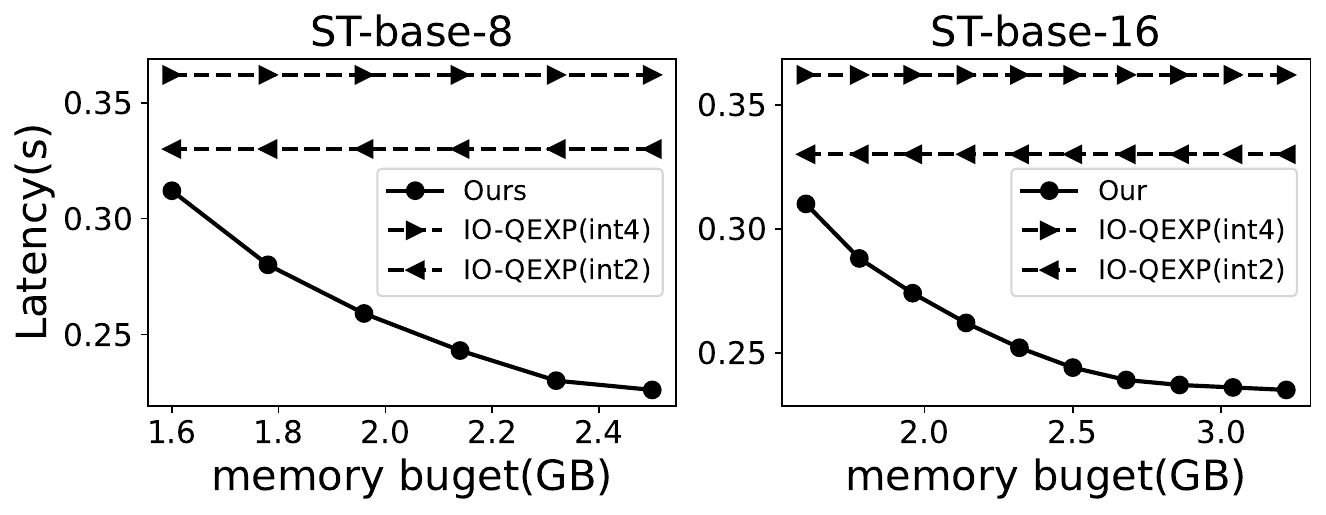}
	}
	\hfil
	\subfloat[Raspberry Pi 4B]{
		\includegraphics[width=0.45\textwidth]{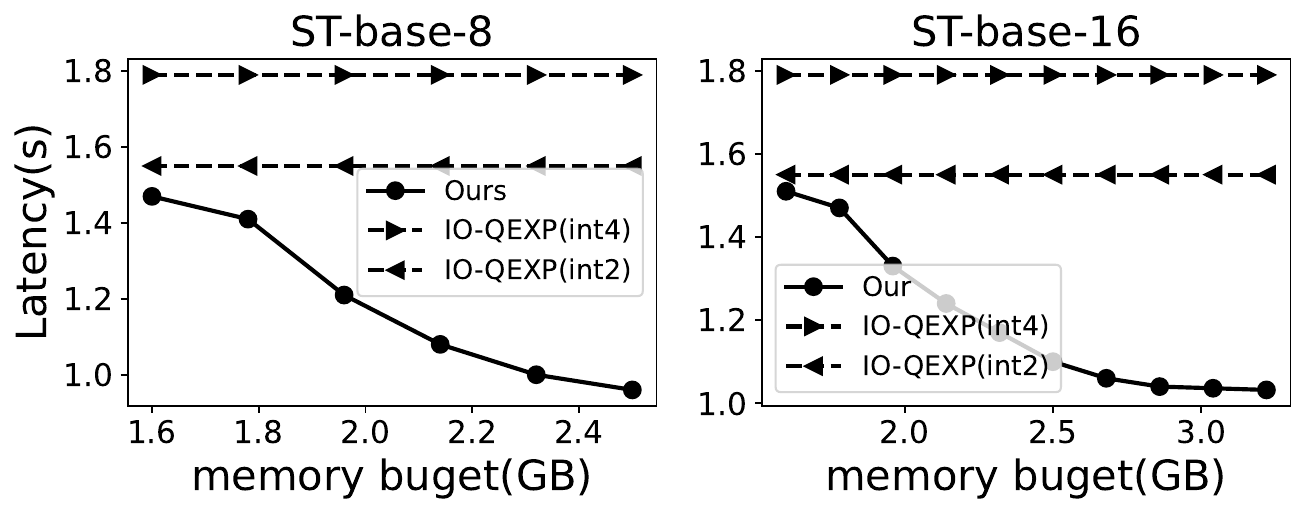}
	}

	\caption{Per-token inference latency(TPOT) of \sys and baselines under different memory budgets.
	}
	\label{fig:eval-cachesize}
\end{figure}
\begin{table}[h!]
    \centering
    \subfloat[ST-base-16]{
        \resizebox{0.24\textwidth}{!}{%
            \begin{tabular}{l|cc}  
                \hline
                Memory(GB)          & Experts (Total 96)   \\ \hline
                1.8         &11 \\ \hline
                2.0         &22  \\ \hline
                2.2         &33  \\ \hline
                3.2         &96  \\ \hline
            \end{tabular} 
        }
    }
    \subfloat[MiniCPM MoE 8x2B (q4)]{
        \resizebox{0.24\textwidth}{!}{%
            \begin{tabular}{l|cc}  
                \hline
                 Memory(GB)          & Experts (Total 320)   \\ \hline
                4.0         &40 \\ \hline
                5.0         &80  \\ \hline
                6.0         &120  \\ \hline
                11.0         &320  \\ \hline
            \end{tabular} 
        }
    }
    \caption{The upper limit of experts that can cached under different memory budgets. 
    The test model of (a) is Switch Transformers base-8, the test device is Jetson TX2, and the non-expert parameters are not quantized. 
    The test model of (b) is MiniCPM MoE 8x2B, the test device is the Xiaomi 14 mobile phone, and the non-expert parameters are quantized to 4 bits.
     }
    \label{tab:eval-cahcemem}
\end{table}

\textbf{Various memory budget.}
\sys adapts to various edge devices with diverse device memory sizes by tailoring the expert buffer, based on the memory budget $M$ ($\S$\ref{sec:design:cache}). 
In our experiments, we configured memory budgets from 1.6GB to 3.5GB, reflecting real-world edge device memory profiles.
We extensively evaluated \sys's per-token inference latency compared to baselines across these memory budgets, and the results are shown in Figure~\ref{fig:eval-cachesize}. 
Notably, as the size of the expert buffer increases, inference latency decreases. 
This is because the expanded expert buffer can retain more previously activated expert weights, leading to higher cache hit ratios and saving weights loading time.

Figure~\ref{fig:eval-cachesize} compares \sys to IO-QEXP on two devices. 
The results consistently show that \sys has lower inference latency across all memory budget configurations compared to both INT4 and INT2 versions of IO-QEXP.

The Table~\ref{tab:eval-cahcemem} shows the upper limit of the number of experts that can be cached under different memory limits. We tested the Swich Transformers base-8 model on Jetson TX2 and the MiniCPM MoE 8x2B model on Xiaomi 14 mobile phone. From the table, we can see that the number of experts that can be cached has a linear relationship with the memory limit. 
The cache buffer will be initialized in the way in $\S$\ref{sec:design-mm:cache}.

\begin{figure}[t]
	\centering					
	\includegraphics[width=0.48\textwidth]{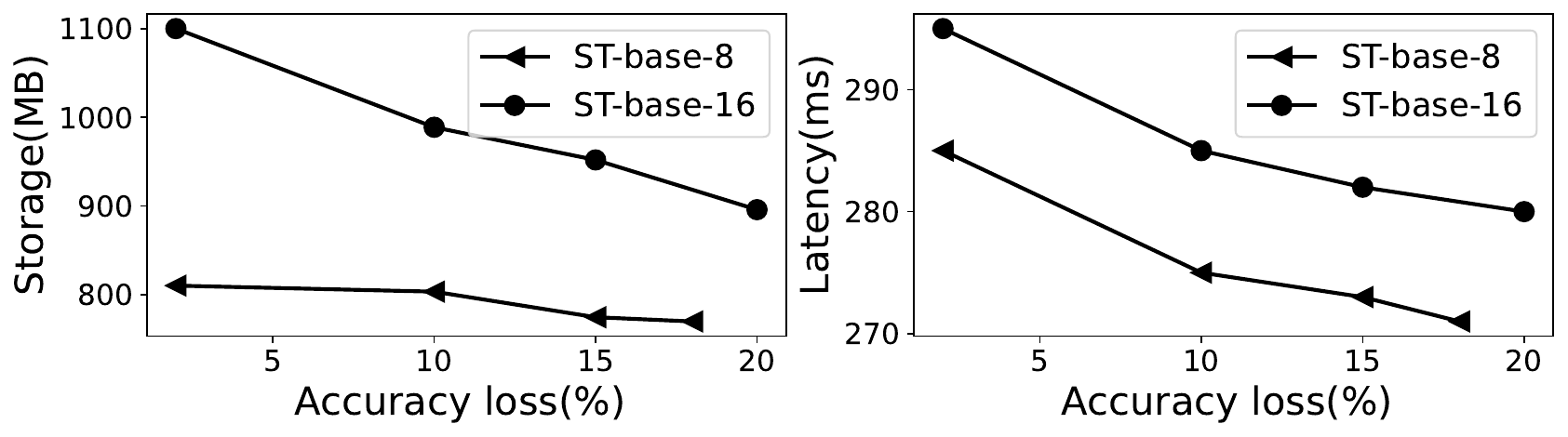}

	\caption{
		The storage and latency of \sys in different desired accuracy loss.
		Models: ST-base-8 and ST-base-16.
		Hardware: Jetson TX2.
	}
	\label{fig:eval-accset}
\end{figure}

\textbf{Impact of tolerable accuracy loss.}
Figure~\ref{fig:eval-accset} provides a comparison across different desired accuracy loss $P$. 
During these experiments, we evaluated inference latency and model storage across accuracy loss levels ranging from 2\% to 20\% on Jetson TX2, running ST-base-8 and ST-base-16 models. 
The results show accuracy loss scales with model sizes and inference latency.
It confirms \sys effectively adapt to available resources (storage and latency) by tuning individual bitwidths of experts. 

\begin{table}[h!]
    \centering
    \subfloat[ST-base-8]{
        \resizebox{0.24\textwidth}{!}{%
            \begin{tabular}{l|cc}  
                \hline
                baseline           &Rouge-2    \\ \hline
                IO-FREE        &0.221 \\ \hline
                Our(Xsum)         &0.220  \\ \hline
                Our(SAMsum)     &0.219   \\ \hline
            \end{tabular} 
        }
    }
    \subfloat[ST-base-16]{
        \resizebox{0.24\textwidth}{!}{%
            \begin{tabular}{l|cc}  
                \hline
                baseline           &Rouge-2    \\ \hline
                IO-FREE        &0.245 \\ \hline
                Our(Xsum)         &0.243  \\ \hline
                Our(SAMsum)     &0.240   \\ \hline
            \end{tabular} 
        }
    }
    
    \subfloat[MiniCPM MoE 8x2B]{
        \resizebox{0.45\textwidth}{!}{%
            \begin{tabular}{l|cccc}  
                \hline
                baseline  &Winograde &Helloswag &MMLU &PiQA    \\ \hline
                IO-FREE   &66.0 &75.1 &50.2 &77.3\\ \hline
                Our(refinedweb) &65.8 &74.8 &49.1 &77,2 \\ \hline
                Our(c4) &64.8 &74.3 &48.9 &77.1  \\ \hline
            \end{tabular}       
        }
    }
    
    \caption{(a) and (b) show the accuracy (Rouge-2) of \sys and IO-FREE test Switch tramsformers base-8 and base-16 models in Xsum dataset
     with different offline dataset.
     (c) shows the accuracy of \sys and IO-FREE test MiniCPM MoE 8x2B model in Winograde, Helloswag, MMLU and PiQA dataset
     with different offline dataset.
     "Our(Xsum)" means the dataset used in the offline stage is Xsum.
     }
    \label{tab:eval-datasetsacc}
\end{table}

\textbf{Impact of the dataset used in the offline stage.}
We discussed the influence of the selected dataset in the offline stage on the model performance. 
We conducted experiments on the Switch tramsformers base-8 and base-16 models, using 300 instances of Xsum or SAMsum dataset in the offline stage. 
And we tested it on Xsum datasets. 
We also conducted experiments on the MiniCPM MoE 8x2B model, with refinedweb or c4 dataset.  
The results are shown in the table~\ref{tab:eval-datasetsacc}. 
It can be seen that compared to the original model's indicators, \sys has almost average 2\% loss for accuracy. 

\begin{figure}[t]
	\centering					
	\subfloat[TPOT (Jetson TX2)]{
		\includegraphics[width=0.47\textwidth]{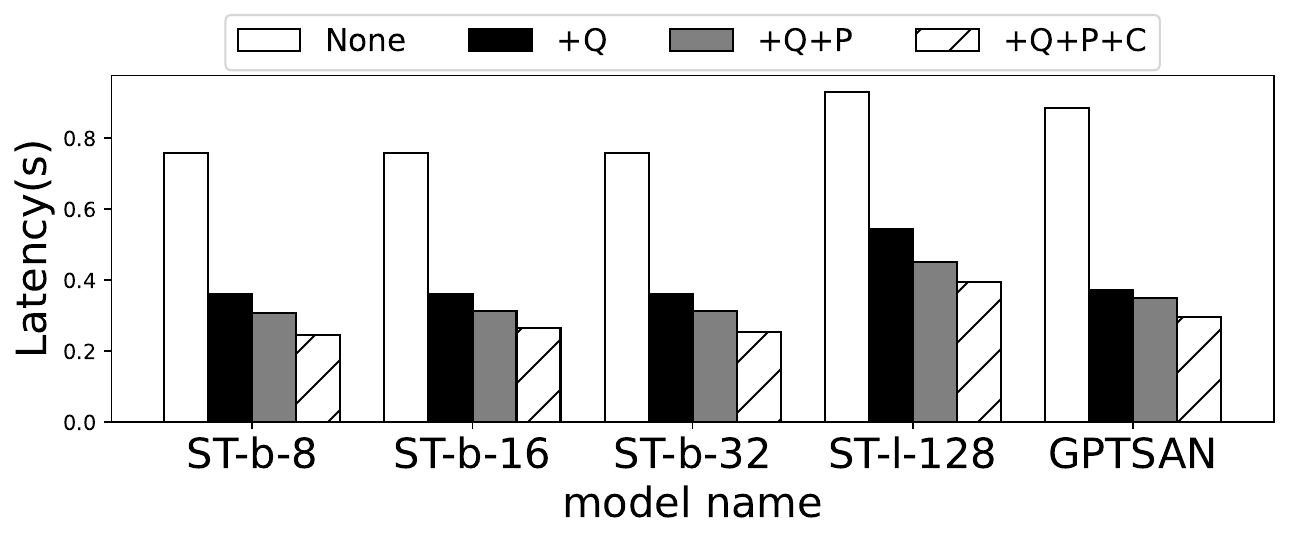}%
	}
	\hfil
	\subfloat[TPOT (Raspberry Pi 4B)]{
		\includegraphics[width=0.47\textwidth]{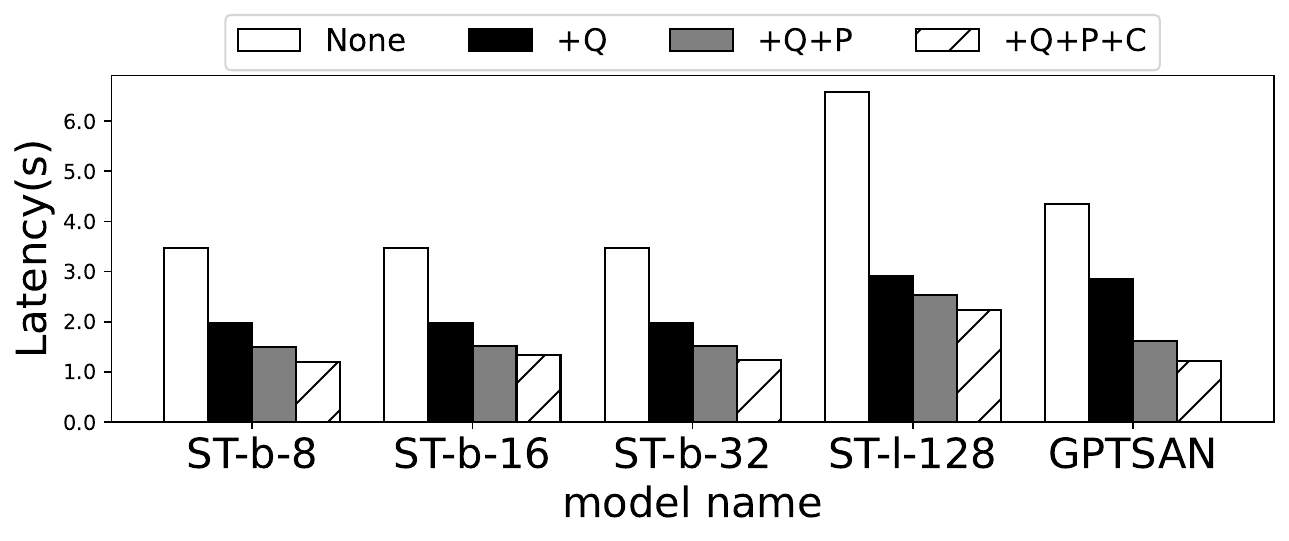}%
	}
	\hfil
	\subfloat[Memory footprint]{
		\includegraphics[width=0.47\textwidth]{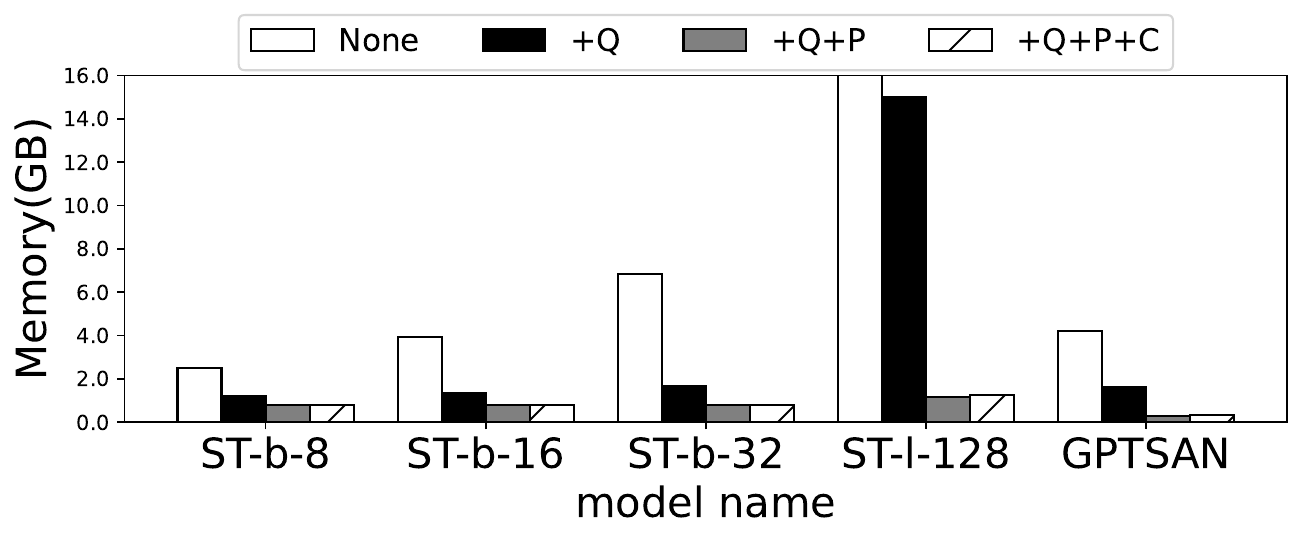}%
	}
	\caption{
	The ablation study results for TPOT and memory footprint of \sys. 
	"Q": expert-wise quantization;
	"P": preloading and pipeline;
	"C": experts buffer.
	}
 \vspace{-10pt}
	\label{fig:eval-ablation}
\end{figure}
\subsection{Ablation Study}
We then evaluate the benefits brought by \sys's each key technique separately. 
The results of per-token inference latency and memory footprint evaluation are illustrated in Figure~\ref{fig:eval-ablation}. 
Our major observation and memory footprint is that each of \sys's key techniques contributes noticeably to the inference speedup.
For example, with ST-base-8 and Jetson TX2, the expert-wise quantization first reduces the inference latency from 0.789s to 0.392s, memory from 2.5GB to 1.21GB.
Preloading and pipeline further reduce the latency to 0.305s, memory to 0.78GB.
Finally, by using expert buffer, the latency finally becomes 0.245s, and memory become 0.81GB.

\section{Related Work}\label{sec:related}

\noindent \textbf{DNN memory optimizations.}
Given that memory is a crucial and scarce resource of mobile devices, memory saving has been an important research direction of the mobile community. 
For instance, Split-CNN~\cite{jin2019split} proposed splitting the weights of a single layer into multiple sub-windows, on which memory offloading and prefetching are applied to reduce activation memory and the weight memory. 
Melon~\cite{wang2022melon} incorporates recomputation and micro-batch to deal with the high memory footprint and fragmentation during on-device training. 
SwapNN~\cite{miao2021enabling} enables large NNs on wimpy MCUs by carefully swapping weights between SRAM and external flash.
\sys shares underlying rationales and techniques with them (quantization and swapping) but exploits unique opportunities of MoE architecture and proposes a novel expert-centric approach to memory saving.

\noindent \textbf{Systems for MoE models.}
Recent researches focused on efficient serving or training MoE-based LLMs~\cite{kong2023serving, sarkar2023edge, du2023sida, frantar2023qmoe}.
For instance, Edge-MoE~\cite{sarkar2023edge} introduces an expert-by-expert computation to maximize the reuse of loaded experts. 
However, these systems are either optimized for distributed clouds or non-autoregressive models such as ViT~\cite{dosovitskiy2020image}.
X-MOE~\cite{chi2022representation} have resolved the representation collapse issue in sparse mixture of experts models.
TA-MoE~\cite{chen2022ta} is a topology-aware routing strategy for large-scale MoE trainging, from a model-system co-design perspective, which can dynamically adjust the MoE dispatch pattern according to the network topology. 
Instead, \sys is a efficient mobile system that exploits unique opportunities to accelerate autoregressive inference of MoE-based LLMs.

\noindent \textbf{Resource-efficient LLMs.}
Research on resource-efficient LLMs has been active. 
Prior works have used methods such as knowledge distillation~\cite{wu2023lamini,li2023symbolic,fu2023specializing,tan2023industry,yuan2023distilling,wang2023scott}, 
network pruning~\cite{frantar2023sparsegpt,ma2023llm,sun2023simple}, 
quantization~\cite{frantar2022gptq,xiao2023smoothquant,lin2023awq,yuan2023rptq,liu2023llm,chee2023quip}, 
architecture design~\cite{liu2023deja,miao2023specinfer,spector2023accelerating,del2023skipdecode,ning2023skeleton,li2023losparse,xu2023tensorgpt}, 
efficient structure design~\cite{dao2022flashattention,dao2023flashattention}, 
and text compression~\cite{valmeekam2023llmzip,chevalier2023adapting,ge2023context} 
to achieve resource-efficient LLMs.
\sys's key designs are orthogonal to those work.

\noindent \textbf{On-device ML optimizations.}
There are two main categories of on-device DNN inference optimizations. 
One is at system level, e.g., by exploiting heterogeneous processors~\cite{cao2017mobirnn,fu2017crnn,huynh2017deepmon,lane2016deepx}, cache~\cite{mathur2017deepeye,xu2018deepcache}, 
generating high-performance GPUs kernels~\cite{liang2022romou}, or adaptive offloading~\cite{laskaridis2020spinn,xu2019deepwear}. 
The other is model level, e.g., quantization~\cite{joo2011fast,liu2018demand} or sparsifiction~\cite{bhattacharya2016sparsification,niu2020patdnn}. 
They reduce the execution time and/or the weights to be read from the disk.
These works can optimize small ML models, but they cannot optimize large language models with running memory that is a hundred times greater than that of edge devices.
\sys is built for resource-efficient MoE-based sparse LLMs and is orthogonal to them.

\section{Conclusions}\label{sec:conclusions}

In this work, we propose \sys, the first on-device inference engine for mixture-of-expert (MoE) LLMs. 
\sys integrates two innovative techniques: expert-specific bitwidth adaptation, reducing expert sizes with acceptable accuracy loss, and expert preloading, which anticipates activated experts and preloads them using a compute-I/O pipeline.
Extensive experiments demonstrate that \sys enables real-time inference for MoE LLMs on edge CPU and GPU platforms while maintaining tolerable accuracy loss.

\section{Acknowledgment}\label{sec:acknowledgment}

This work was supported by NSFC (U21B2016, 62032003, 62425203), Fundamental Research Funds for the Central Universities under Grant 2024ZCJH11.
Liwei Guo was partly supported by Sichuan Science and Technology Plan “Unveiling and Leading” Project (No. 2024YFCY0001).
Shiyun Wei and Ao Zhou are the corresponding authors.
We thank the anonymous reviewers for their valuable suggestions.

\bibliographystyle{plain}
\bibliography{IEEEabrv,bib/ref-mwx}

\appendices

\section{\sys Examples}\label{sec:appendix:example}
This appendix shows the output of the Switch Transformers base-8 model in the original version and after \sys processing. Here are two examples provided:

\begin{mdframed}[backgroundcolor=gray!10]
\begin{itemize}[left=0pt, label={}, noitemsep, topsep=0pt,]
  \item \textbf{\textcolor{blue}{Prompt:}} 
    summarize: The 18-year-old identical twins have come through the club's academy to impress in nine Premiership appearances between them this season.
    Both play in the back row and have also featured for the England Under-20 side.
    "They will play key parts in the club's vision of developing players in the academy, and bringing them through to the first team," Sale director of rugby Steve Diamond said.
    The pair became only the fourth set of twins to play side-by-side in the Premiership when they appeared in Sale's 34-24 defeat by Wasps on 27 November.
    Tom is also the Sharks' youngest Premiership try scorer after crossing on his debut in the 31-13 win over Bristol on 30 October.
    
  \item \textbf{\textcolor{orange}{original:}}
  \begin{itemize}[left=0pt,label={}, noitemsep, topsep=0pt]
  \item Sale's academy has produced nine Premiership players this season. Tom is the Sharks' youngest player.
  \end{itemize}
  
  \item \textbf{\textcolor{orange}{\sys:}}
  \begin{itemize}[left=0pt,label={}, noitemsep, topsep=0pt]
  \item The twins have played in the back row and have also played for the England Under-20 team. They are the fourth set of twins to play in the Premiership this season.
  \end{itemize}
\end{itemize}
\end{mdframed}

\begin{mdframed}[backgroundcolor=gray!10]
\begin{itemize}[left=0pt, label={},noitemsep,topsep=0pt]
  \item \textbf{\textcolor{blue}{Prompt:}} 
    summarize: Wellington monument on the Blackdown Hills, in Somerset, was built in 1817 but since 2005 it has been fenced off because of falling stone debris.
    The National Trust is using ground-penetrating radar on the 174ft (53m) tower to see under its stone cladding.
    Ken Evans, from the trust, said the work was "crucial".
    Built on one of the highest points of the Blackdown Hills, the landmark was put up as a tribute to the Duke of Wellington's military achievements at the Battle of Waterloo.
    But according to the trust, it has been struck by lightning twice in its history and renovating the very tall landmark every 10 to 15 years has been "expensive and unsustainable".
    Mr Evans, the trust's building surveyor, said the radar study was one of several being carried out to "understand this unique and somewhat complex monument".
    "We have been using wind and movement sensors which have already surprised us by showing that it doesn't flex in the wind quite as much as we expected," he said.
    "The ground-penetrating radar seeks to identify voids and gaps in the stonework under the surface but should also tell us more about the materials which were used to build the obelisk."
    Data from the detailed survey will also be used to build a computer model of the obelisk and help with a "more effective repair approach".
  \begin{itemize}[left=0pt,label={},noitemsep,topsep=0pt]
  \item
  \end{itemize}
    
  \item \textbf{\textcolor{orange}{original:}}
  \begin{itemize}[left=0pt,label={},noitemsep,topsep=0pt]
  \item The National Trust is studying Wellington monument on the Blackdown Hills, Somerset. The monument has been struck by lightning twice in its history. The radar study is one of several studies carried out to understand the monument.
  \end{itemize}
  
  \item \textbf{\textcolor{orange}{\sys:}}
  \begin{itemize}[left=0pt,label={},noitemsep,topsep=0pt]
  \item The National Trust is working on a project to repair Wellington monument on the Blackdown Hills, Somerset. The monument has been struck by lightning twice in its history. The project is important for the trust's building surveyor. The project will help the National Trust with the repair approach.
  \end{itemize}
\end{itemize}
\end{mdframed}


 

\begin{IEEEbiography}[{\includegraphics[width=1in,height=1.25in,clip,keepaspectratio]{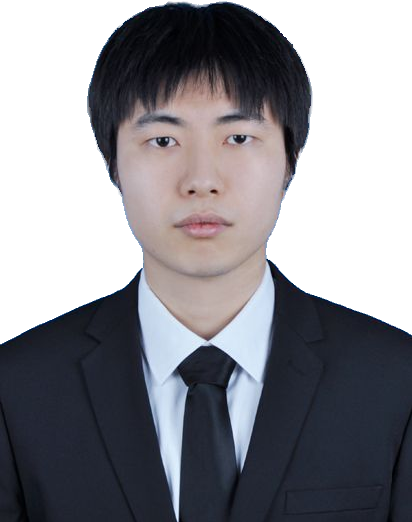}}]{Rongjie Yi} is a Ph.D. student at the School of Computer Science, Beijing University of Posts and Telecommunications, China. 
\end{IEEEbiography}


\begin{IEEEbiography}[{\includegraphics[width=1in,height=1.25in,clip,keepaspectratio]{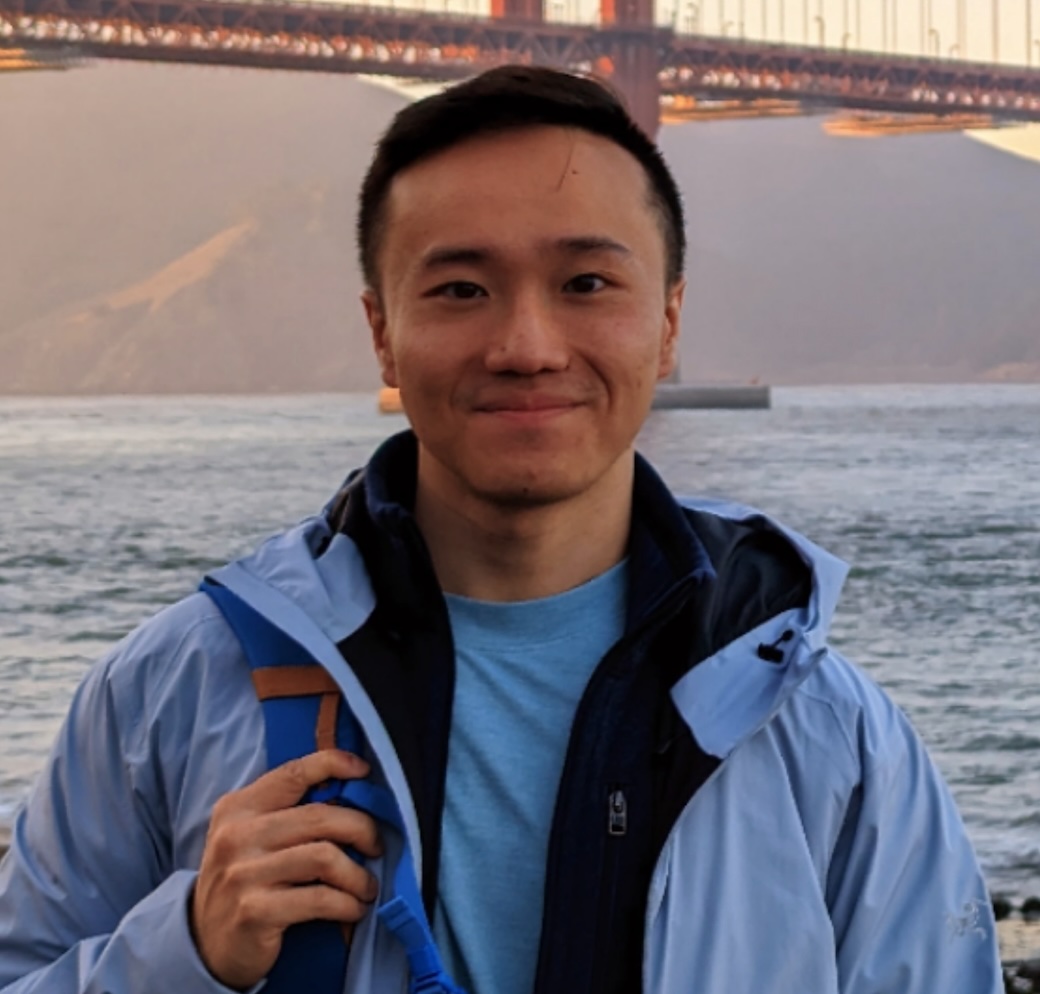}}]{Liwei Guo} is a tenure-track Assistant Professor at the University of Electronic Science and Technology of China (UESTC) in the school of Computer Science. 
He received his Ph.D. degree from the University of Virginia in 2022 under Prof. Felix Xiaozhu Lin. 
He is interested in improving the efficiency and security of edge devices from the perspective of systems software. 
For more details, please visit \url{https://zaxguo.github.io}.
\end{IEEEbiography}

\begin{IEEEbiography}[{\includegraphics[width=1in,height=1.25in,clip,keepaspectratio]{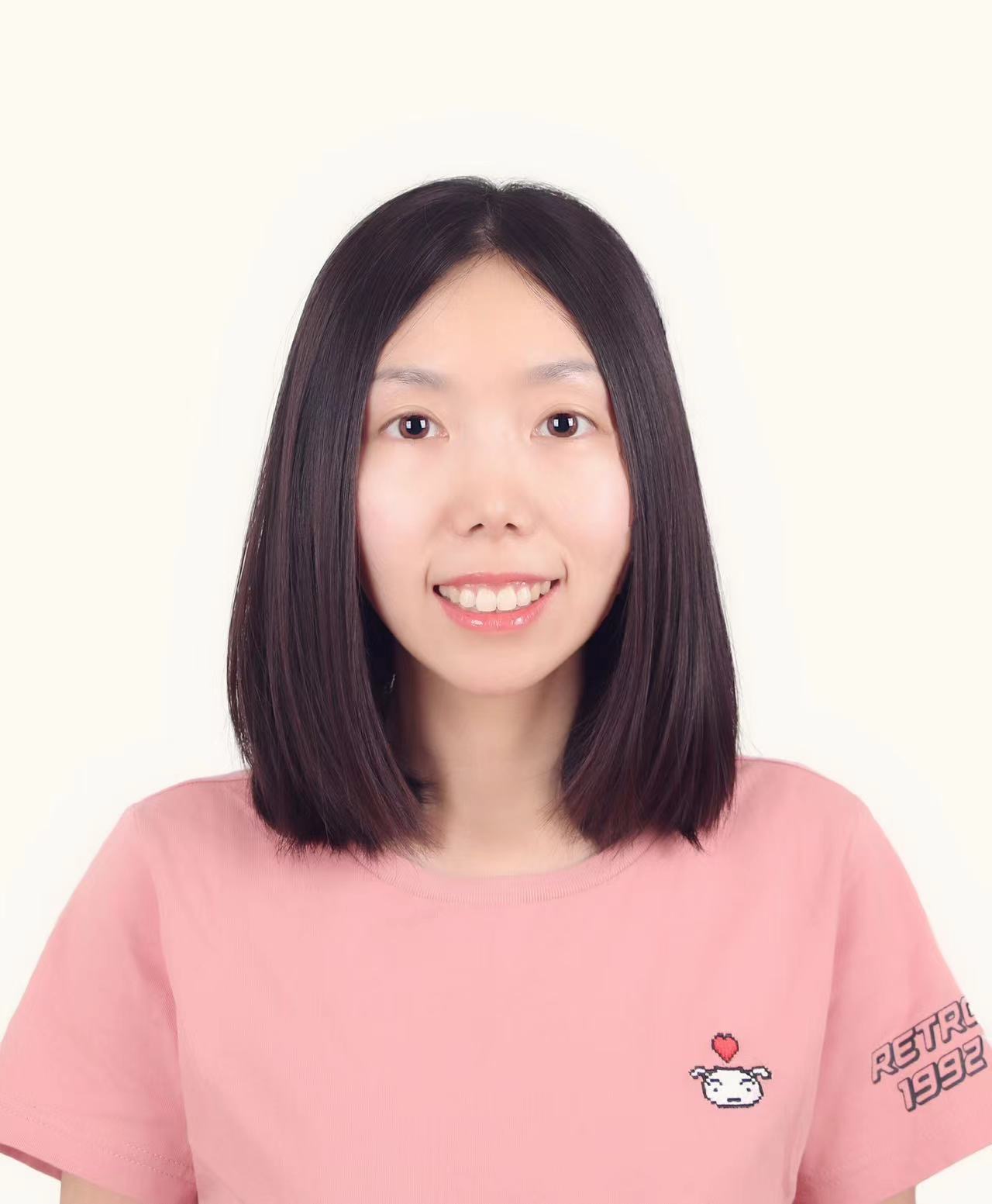}}]{Shiyun Wei} is an engineer in the Zhongguangcun Laboratory. Her research interests include program analysis and software engineering.
\end{IEEEbiography}

\begin{IEEEbiography}[{\includegraphics[width=1in,height=1.25in,clip,keepaspectratio]{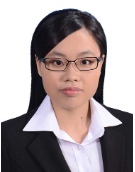}}]{Ao Zhou} received the Ph.D. degrees from the Beijing University of Posts and Telecommunications, Beijing, China, in 2015. She is currently an associate professor with the State Key Laboratory of Networking and Switching Technology, Beijing University of Posts and Telecommunications. She has published more than 20 research papers. She played a key role at many international conferences. Her research interests include cloud computing and edge computing.
\end{IEEEbiography}

\begin{IEEEbiography}[{\includegraphics[width=1in,height=1.25in,clip,keepaspectratio]{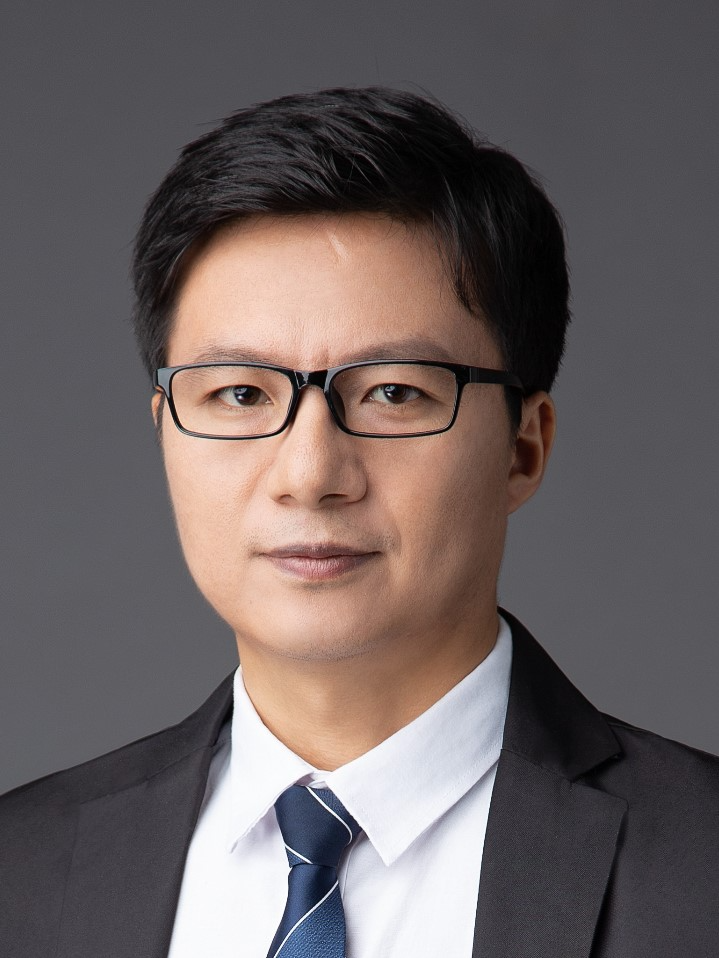}}]{Shangguang Wang} is a Professor at the School of Computer Science, Beijing University of Posts and Telecommunications, China. He received his Ph.D. degree at Beijing University of Posts and Telecommunications in 2011. He has published more than 150 papers. His research interests include service computing, mobile edge computing, and satellite computing. He is currently serving as Chair of IEEE Technical Committee on Services Computing, and Vice-Chair of IEEE Technical Committee on Cloud Computing (2020-). He also served as General Chairs or Program Chairs of 10+ IEEE conferences. He is a Fellow of the IET, and Senior Member of the IEEE. For further information on Dr. Wang, please visit: \url{http://www.sguangwang.com}.
\end{IEEEbiography}

\begin{IEEEbiography}[{\includegraphics[width=1in,height=1.25in,clip,keepaspectratio]{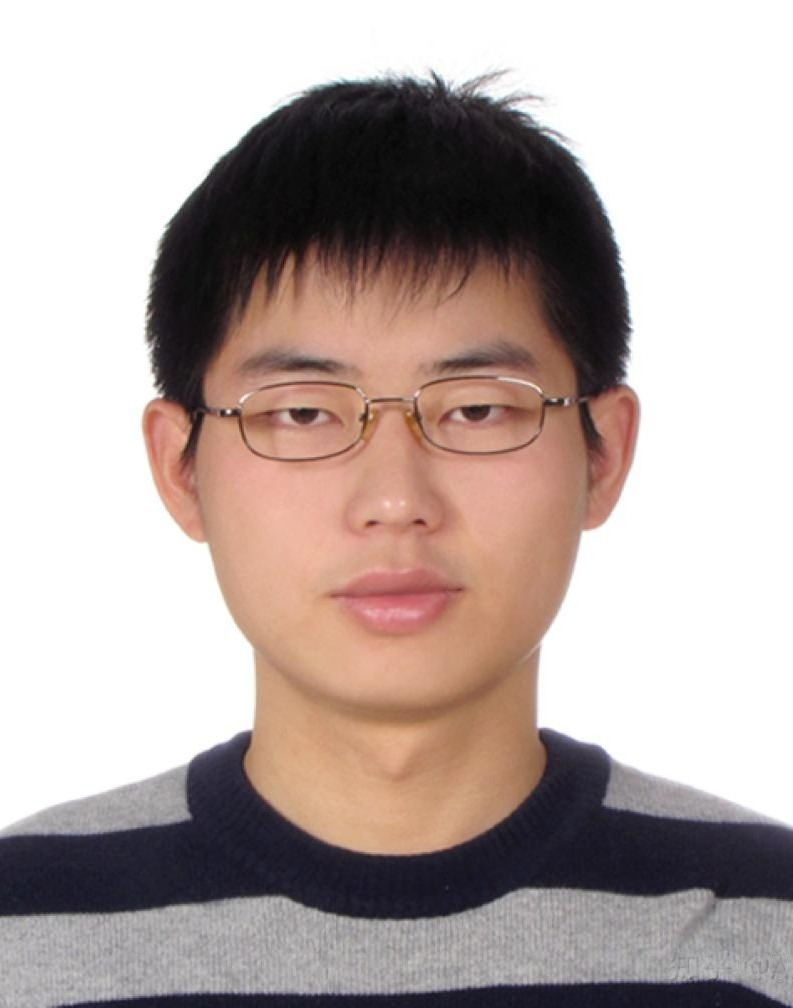}}]{Mengwei Xu} is an associate professor in the computer science department at Beijing University of Posts and Telecommunications. His research interests cover the broad areas of mobile computing, edge computing, artificial intelligence, and system software.
\end{IEEEbiography}


\vfill

\end{document}